\documentclass[preprint,11pt,3p]{elsarticle}

\usepackage{graphicx}
\usepackage{tabularx}
\usepackage{amssymb}
\usepackage{paralist}
\usepackage{amsmath,mathtools}
\usepackage{makecell}
\usepackage{enumitem}
\usepackage{float}
\usepackage{hyperref}
\usepackage{multirow}
\usepackage[table]{xcolor}
\usepackage{pifont} 
\usepackage{tikz}
\usepackage{csquotes}
\biboptions{sort&compress}
\usepackage{rotating}
\usepackage{booktabs} 
\usepackage{array}    
\usepackage{longtable}

\usepackage{url}
\usepackage{doi}
\usepackage{xcolor}
\definecolor{mygreen}{rgb}{0,0.6,0}
\usepackage{subcaption}
\usepackage{bbm}
\usepackage{threeparttable}

\usepackage{nomencl}
\usepackage{arydshln} 
\usepackage{algorithmicx,algorithm,algpseudocode}
\usepackage{siunitx}

\algdef{SE}[DOWHILE]{Do}{doWhile}{\algorithmicdo}[1]{\algorithmicwhile\ #1}%

\algrenewcommand\algorithmicrequire{\textbf{Input:}}
\algrenewcommand\algorithmicensure{\textbf{Output:}}

\makenomenclature

\newcommand{\cyes}{\ding{51}}

\allowdisplaybreaks

\hypersetup{
	colorlinks   = true, 
	urlcolor     = blue, 
	linkcolor    = blue, 
	citecolor    = blue 
}

\begin{document}

\begin{frontmatter}


\title{SCOPE: A Lightweight-training LLM Framework for Air Traffic Control Readback Monitoring}



\author[hkust]{Qihan Deng}
\author[buaa,sklab]{Minghua Zhang}
\author[buaa,sklab]{Yang Yang\corref{cor1}}
\author[hkust]{Zhenyu Gao\corref{cor1}}
\cortext[cor1]{Corresponding authors.}
\address[hkust]{Department of Mechanical and Aerospace Engineering, The Hong Kong University of Science and Technology, Clear Water Bay, 999077, Hong Kong}
\address[buaa]{School of Electronic and Information Engineering, Beihang University, Beijing, 100191, China}
\address[sklab]{State Key Laboratory of CNS/ATM, Beijing, 100191, China}

\begin{abstract}

Pilot readback of Air Traffic Control (ATC) voice instructions is a primary safeguard against miscommunication in air transportation. However, readback anomalies remain implicated in approximately 80\% of aviation incidents. This vulnerability is further exacerbated by rising traffic volume and elevated cognitive workload, thereby motivating automated readback monitoring by machine. Traditional rule-based and machine learning approaches struggle to generalize across the highly variable and evolving phraseology of air traffic controller--pilot communications. While Large Language Models (LLMs) have opened a new avenue through their strong reasoning and generalization capabilities, existing approaches still face deployment and computational barriers in practice. In this work, we propose Semantic reasoning for Communication via Open-set Plug-in with Examples (SCOPE), a novel lightweight-training LLM framework that advances both the efficiency and accuracy of machine-based ATC readback monitoring. The core idea is to couple a plug-in open-set classifier with a carefully designed in-context learning mechanism on top of a frozen LLM. Extensive experiments on the semi-synthetic communication dataset show that SCOPE attains superior accuracy while delivering the low-latency response required for operational environments. Under a few-shot setting, SCOPE achieves 91.05\% accuracy in open-set detection and corrects 96.63\% of anomalous readbacks, thereby outperforming the strongest available baselines while providing explanations for its decisions. These findings demonstrate the potential of our framework as a practical pathway toward interpretable and controllable ATC readback monitoring.

\end{abstract}

\begin{keyword}
Air traffic control \sep Air traffic controller--pilot communication \sep Readback monitoring \sep In-context learning \sep Large language model
\end{keyword}

\end{frontmatter}



\section{Introduction}

The reliable delivery of Air Traffic Control (ATC) instructions is essential for safe and efficient aviation operation~\citep{icao_doc4444}. As a human-in-the-loop supervisory control system, ATC relies primarily on voice radio communication between Air Traffic Controllers (ATCos) and pilots, whereby ATCos issue instructions and pilots carry out the corresponding operations~\citep{icao_doc9432}. With the continued growth of air traffic volume, ATCo workload has risen considerably, increasing the likelihood of communication errors~\citep{goppel2026operationalizing}. According to a survey by the National Aeronautics and Space Administration (NASA), miscommunication serves as either a causal or circumstantial factor in approximately 80\% of aviation incidents and accidents~\citep{skybrary_pilot_controller}. The traditional safeguard against such miscommunication is the readback protocol, in which pilots verbally repeat ATCo instructions to confirm their reception. However, this procedure depends entirely on human attention and memory, which could become unreliable under high-density operations and elevated cognitive workload. Recent accidents have exposed this vulnerability in concrete operational terms, from the misinterpretation of ATC departure sequencing as runway entry clearance in the 2024 Haneda runway collision to the incomplete information in the 2025 midair collision near Ronald Reagan Washington National Airport~\citep{YAN2025106913,NTSB2026DCA}. Such incidents motivate automated readback monitoring as a technological complement to the readback protocol.

Automated readback monitoring checks whether a pilot's readback is consistent with the preceding ATCo instruction, detecting inconsistencies and providing early warnings for potential communication anomalies overlooked by human ATCos. Early studies pioneered readback error detection by Automatic Speech Recognition (ASR) but relied on hand-crafted rules and templates that generalize poorly to highly variable ATC phraseology~\citep{chen2017read}. Subsequent work began to leverage Machine Learning (ML) approaches, such as reformulating readback monitoring as a sequence-level semantic matching task using Long Short-Term Memory (LSTM) encoders~\citep{lu2016new,jia2017verification}, replacing manual rules with learned representations. \citet{guimin2022novel} further introduced deeper attention-based layers that captured token-level matching. Beyond single-task methods, \citet{lin2019real} embedded readback monitoring into a broader ATC safety system, while the HAAWAII project~\citep{helmke2021readback,helmke2022readback} identified mismatches between ATCo instructions and pilot readbacks in Icelandic en-route operations. Despite these advances, existing ML approaches remain limited by narrow, task-specific designs and lack the broad semantic understanding required to generalize to the diverse and evolving communications encountered in real-world ATC operations.

The emergence of Large Language Models (LLMs) has opened a transformative new direction for readback monitoring through their strong contextual reasoning and broad linguistic generalization capabilities~\citep{brown2020language}. In the current literature, \citet{connolly2024aircraft} provided an early proof of concept by adapting LLMs to aviation anomaly detection, while \citet{SEMIHSADAK2026132241} decomposed error detection into four specialized LLM agents fused with safety rules through Bayesian inference. Nevertheless, existing LLM-based approaches for readback monitoring still face practical barriers to ATC deployment. For instance, task adaptation and inference costs introduce substantial overhead that is difficult to reconcile with real-time operational requirements. Therefore, the challenge lies in developing a framework that satisfies both the safety-critical and real-time demands of operational ATC environments.

\begin{figure}[h!]
\centering
\includegraphics[width=0.95\linewidth]{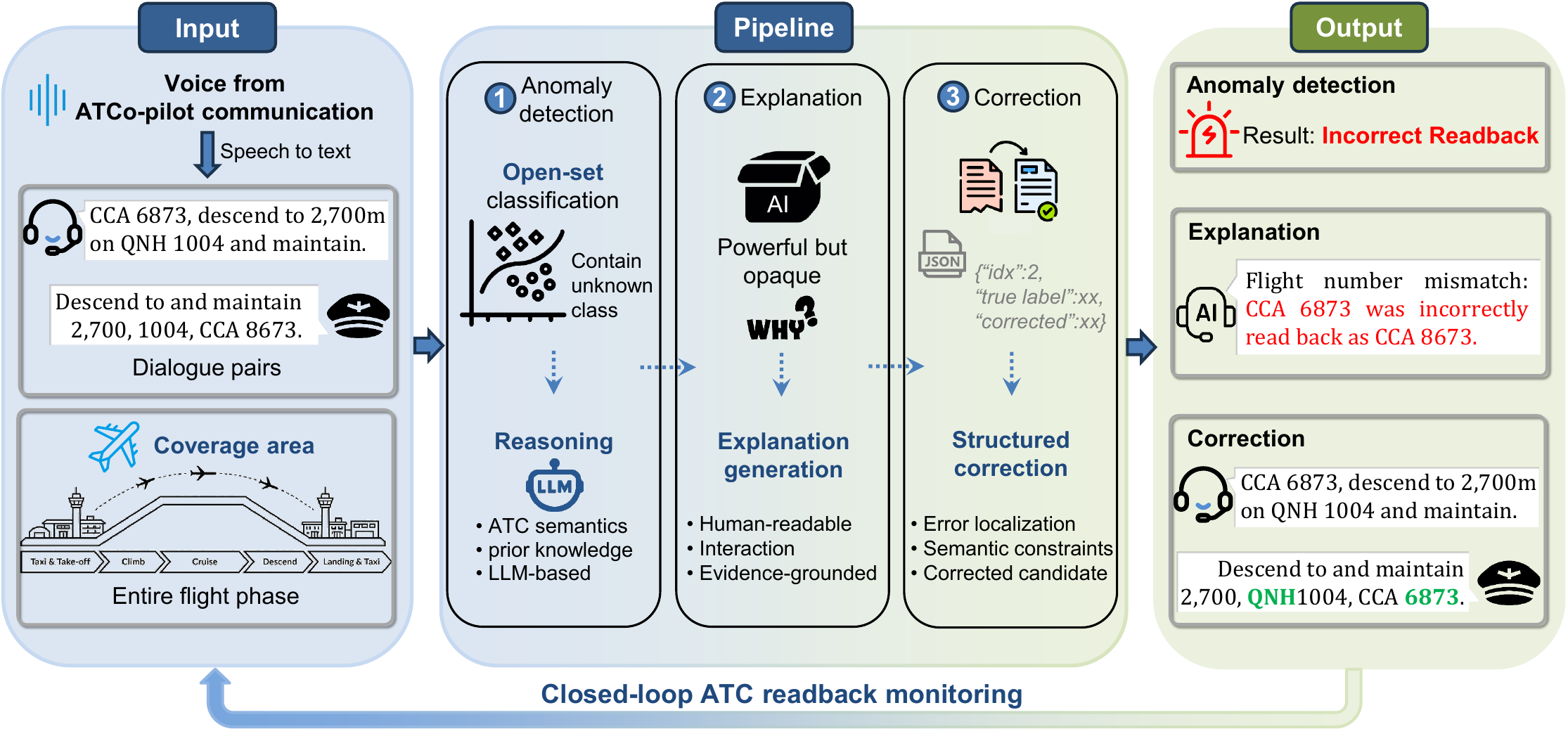}
\caption{Overview of the proposed closed-loop ATCo--pilot communication readback monitoring pipeline.}
\label{fig:pipeline}
\end{figure}

To better support ATCo--pilot communication monitoring, this paper proposes \textit{Semantic reasoning for Communication via Open-set Plug-in with Examples} (SCOPE), a lightweight-training LLM framework for {\bf accurate} and {\bf efficient} ATC readback monitoring. As illustrated in \hyperref[fig:pipeline]{Fig. 1}, SCOPE takes ATCo instructions and corresponding pilot readbacks as input and produces three cascaded outputs: \underline{anomaly detection}---detection results that identify anomalous or unknown readbacks; \underline{explanation}---natural language explanations that articulate the semantic source of any mismatch; and \underline{correction}---standardized correction suggestions. To achieve both accuracy and efficiency in LLM-based ATC readback monitoring, the proposed SCOPE framework is built upon (1) a Plug-in Open-set Classifier (POC) and (2) an In-Context Learning (ICL) mechanism that uses a Diverse Example instruction Anchored Retrieval (DEAR) strategy and an Air Traffic Chain-of-Thought (ATCoT) reasoning mechanism. Specifically, POC adds open-set awareness to ICL by coupling a lightweight classifier with a frozen LLM. It supplies stable label priors for known readback states and identifies unknown communications using independent boundaries and K-Nearest Neighbor (KNN) scoring. DEAR enhances ICL by anchoring retrieval on ATCo instructions for scenario relevance while selecting diverse intra-class readback examples to enrich context. ATCoT further bolsters semantic reasoning, directing the LLM to construct and compare intents and critical slots before final prediction. Together, these components allow SCOPE to advance state-of-the-art accuracy in ATC readback detection without inheriting the efficiency and cost overheads of prior LLM approaches, thus achieving the low inference cost and low latency required for real operational deployment.

The contributions of this work are summarized as follows:
\begin{itemize}
\item We propose SCOPE, an end-to-end ATCo–pilot communication monitoring system that unifies readback anomaly detection, error explanation, and standardized correction.
\item We introduce a lightweight-training paradigm that leverages a locally deployed LLM through ICL. Within this paradigm, three coordinated modules equip the LLM with open-set recognition, scenario-aware example retrieval, and structured ATC knowledge reasoning over readback content.
\item We conduct extensive experiments on the operationally grounded communications dataset and demonstrate that SCOPE achieves superior accuracy while supporting low-latency response, paving the way for LLM-based automation in ATC operations.
\end{itemize}


\section{Related Work}\label{sec:work}

\subsection{ATCo--Pilot Communication Readback Monitoring}

Early efforts to improve ATCo--pilot communication safety mainly focused on standardized procedures, improved operational environments, and better supporting systems~\citep{icao_doc4444,icao_doc9835}. These measures reduce the chance of miscommunication, but their effectiveness still depends on ATCo attention, which is inherently limited under high workload conditions. Such limitations become particularly pronounced during dense information transmissions or frequent exchanges across different languages~\citep{molesworth2015miscommunication,wu2019investigation}. The preventive measures above cannot determine whether the content of an instruction has been correctly delivered and acknowledged. This gap motivated the shift toward automated readback safety monitoring, which has since progressed along three intertwined threads: dataset construction, methodological development, and operational deployment.

Dataset construction has provided the empirical foundation for data-driven readback monitoring and instruction understanding. The ATCO2 dataset released over 5{,}000 hours of ATCo--pilot communications across more than ten airports with partial annotations for ASR, role detection, and named entity recognition~\citep{zuluaga2022atco2,zuluaga2023lessons}. ATCSpeech and its extensions provided multilingual ATC corpora enriched along accent, channel noise, and English-Mandarin code-switching dimensions~\citep{yang2019atcspeech,lin2020unified,lin2021atcspeechnet}. The more recent ATSIU benchmark contributed 19.8k transcribed utterances with 9 coarse and 26 fine intents plus 78 slots, supporting instruction understanding across all flight phases~\citep{zhang2025atsiu}. Together, these datasets have enabled the training and comparison of data-driven readback monitoring models.

Building on these datasets, readback monitoring was initially formulated as semantic matching, with LSTM and convolutional neural network encoders replacing handcrafted rules through learned sentence representations~\citep{lu2016new,jia2017verification,cheng2018readback}. \citet{guimin2022novel} moved beyond sentence-level limitations by introducing token-level interaction in an attention-based BiLSTM. Subsequent work pushed toward deeper semantic understanding through knowledge-augmented Bidirectional Encoder Representations from Transformers (BERT) for few-shot ATC intent recognition~\citep{HUI2025113524} and through hierarchical multi-task learning for joint slot filling, role detection, and intent recognition in ATCo--pilot communication~\citep{ZHANG2026103812}. Related open-world modeling has also begun to appear in broader aviation safety applications. \citet{yang2024scenario} proposed a multimodal Transformer for air crisis event recognition, jointly handling known event classes and unknown aviation crisis events. Nevertheless, these methods either treat the task as a closed-set classification problem or stop at the instruction understanding stage, leaving operational readback verification insufficiently addressed.

The third line of work focuses on operational deployment, aiming to integrate ATCo--pilot communication monitoring into real safety support systems. The tower prototype by \citet{chen2017read} established the early feasibility of readback checking for runway incursion prevention, while the HAAWAII project demonstrated en-route deployment at operational scale and exposed the inherent trade-off between detection coverage and false alarm tolerance~\citep{helmke2021readback,helmke2022readback}. Complementary work further showed that readback monitoring becomes more effective when embedded in a broader safety loop incorporating flight status information~\citep{lin2019real}. Other studies showed that automatic readback generation can reduce reliance on pseudo pilots in training simulators~\citep{zhang2022repetition}, and that timely error detection can lead to measurable risk reduction in both loss of en-route separation~\citep{sun2021automatic} and surface movement collisions~\citep{pang2026voice}.

As LLMs continue to advance in both capability and reliability, a growing body of work has begun to explore their use in ATC applications. Domain adaptation studies have considered training pretrained transformers~\citep{nielsen2024towards}, self-supervised pipelines paired with LLM-as-a-Judge evaluation~\citep{ge2025aviation}, and continual pre-training that injects aviation knowledge from flight documentation~\citep{zhang2026aviationcopilot}. \citet{connolly2024aircraft} investigated whether GPT-4 could generalize to aviation anomaly reasoning through prompt-based ATC event detection, and \citet{SEMIHSADAK2026132241} decomposed error detection into four specialized LLM agents combined using Bayesian rules. Yet hallucination, detection precision, and inference timeliness remain persistent obstacles to safety-critical deployment.

\begin{table}[ht]
\centering
\caption{Comparison of related work. Compared with existing studies, SCOPE enables ATC understanding and response under open-set conditions, achieving both zero-shot and few-shot performance without LLM training.}
\label{tab:related_work}
\setlength{\tabcolsep}{5pt}
\fontsize{8}{10}\selectfont
\begin{tabular}{lcccccc}
    \toprule
    \multirow{2}{*}{\textbf{Study}}
        & \multicolumn{2}{c}{\textbf{Scenario}}
        & \multicolumn{2}{c}{\textbf{Experimental Setup}}
        & \multicolumn{2}{c}{\textbf{Capability}} \\
    \cmidrule(lr){2-3} \cmidrule(lr){4-5} \cmidrule(lr){6-7}
        & Dataset & Open-Set & Training-Free & Zero/Few-Shot
        & Understanding & Response \\
    \midrule
    \rowcolor{gray!15} \shortstack[l]{DeepSpeech2 \\ \citep{amodei2016deep}} & LibriSpeech & &  & & \cyes &  \\[4pt]
    \rowcolor{gray!15} \shortstack[l]{BERT \\ \citep{DBLP:conf/naacl/DevlinCLT19}} & ATIS/SNIPS & &  & & \cyes & \\[4pt]
    \rowcolor{gray!15} \shortstack[l]{ICL \\ \citep{brown2020language}} & SuperGLUE &  & \cyes & \cyes & \cyes & \\[4pt]
    \rowcolor{gray!15} \shortstack[l]{MilTOOD \\ \citep{DarrinSGCPC24}} & MilTOOD-C & \cyes &  & & \cyes &  \\
    \midrule
     \shortstack[l]{ATCSpeechNet \\ \citep{lin2021atcspeechnet}}& ATCSpeech &  &  & & \cyes & \\[4pt]
     \shortstack[l]{TIU-RIG \\ \citep{zhang2022repetition}}& ATCC-China &  &  & & \cyes & \cyes\\[4pt]
     \shortstack[l]{SRD \\ \citep{zuluaga2023lessons}}  & ATCO2 &  &  & & \cyes & \\[4pt]
     \shortstack[l]{STUEC \\ \citep{yang2024scenario}}  & AirCrisisMMD & \cyes &  & & \cyes & \\[4pt]
    \shortstack[l]{KaFIR \\ \citep{HUI2025113524}} & ATSIU & &  & \cyes & \cyes & \\[4pt]
    \shortstack[l]{ASSC-AI \\ \citep{pang2026voice}} & LiveATC & &  &  & \cyes & \cyes\\[4pt]
    \shortstack[l]{MTHN \\ \citep{ZHANG2026103812}} & ATSIU & &  &  & \cyes & \\[4pt]
    \shortstack[l]{CSMA \\ \citep{SEMIHSADAK2026132241}} & UWB-ATCC & &  &  & \cyes & \\[4pt]
    \hdashline\addlinespace[2pt]
    \textbf{SCOPE (Ours)} & APCP & \cyes & \cyes & \cyes & \cyes & \cyes \\
    \bottomrule
\end{tabular}

\vspace{2pt}
\raggedright
\footnotesize{\textit{Note:} \ding{51} indicates the capability is supported. 
\colorbox{gray!15}{Gray shading} indicates methods from the general NLP domain.}
\end{table}

\subsection{General Natural Language Processing}
Research on ATCo--pilot communication understanding builds on a broader foundation of general Natural Language Processing (NLP) techniques, including standard benchmarks for speech and language understanding, methods for open-set recognition, and the recent shift from full training toward effective adaptation with LLMs.

Large annotated benchmarks have long served as the proving ground for speech and language understanding. LibriSpeech~\citep{panayotov2015librispeech} remains a standard thousand-hour corpus for ASR and has been widely adopted by end-to-end models such as DeepSpeech2~\citep{amodei2016deep}. ATIS~\citep{DBLP:conf/naacl/HemphillGD90} and SNIPS~\citep{DBLP:journals/corr/abs-1805-10190} serve as the benchmarks for intent recognition and slot filling, the two core tasks of language understanding that follow ASR in the standard NLP pipeline. Built upon these resources, pretrained transformer encoders such as BERT~\citep{DBLP:conf/naacl/DevlinCLT19} and its advanced variant DeBERTa~\citep{he2021deberta} provide the widely used backbones for downstream classification and sequence labeling.

\textbf{Open-set recognition.} Standard classifiers assume that all test samples belong to classes seen during training, but real-world systems often encounter inputs that do not fit any known class and should be detected rather than forcibly classified. Early approaches tackled this problem by recalibrating classifier outputs. DOC~\citep{DBLP:conf/emnlp/ShuXL17} replaces the softmax layer with independent sigmoid heads and learns a detection threshold for each class, while OpenMax~\citep{bendale2016towards} reshapes the output scores using statistics of training activations to expose inputs that deviate from typical patterns. A simpler baseline, MSP~\citep{hendrycks2016baseline}, shows that the confidence score of a standard classifier alone is already a useful signal for detecting unknown inputs. Later methods extend this line of work in two directions. Outlier exposure~\citep{DBLP:conf/iclr/HendrycksMD19} introduces auxiliary unknown samples, so that the model learns to separate known from unknown inputs rather than relying solely on inference-time heuristics. Representation-based methods instead extract detection signals from internal layers of the encoder. MilTOOD~\citep{DarrinSGCPC24} observes that the final layer is often not the most discriminative for unknown detection, and aggregates similarity-based anomaly scores across all layers, outperforming baselines that rely only on the last layer. These techniques offer practical tools for handling the open-set nature of ATC communication, where non-readback utterances need to be reliably detected.

\textbf{From parameter updating to parameter-free adaptation.} The dominant paradigm for adapting pretrained models to new tasks has long been full or parameter-efficient training. Modern LLMs such as ChatGPT~\citep{openai2026gpt53instantcard}, the Qwen series~\citep{yang2025qwen3}, and the Phi series~\citep{abdin2024phi} have pushed this paradigm to new scales, combining massive pretraining with supervised and reinforcement-based alignment to reach strong performance across diverse tasks. As these models grew, ICL emerged as a new capability, in which a frozen model adapts to a new task purely from a handful of examples placed in the prompt, without any gradient updates~\citep{brown2020language}. Because downstream performance under ICL depends heavily on which examples are selected, recent work has focused on the retrieval and construction of effective example pools. DICL~\citep{KapuriyaKGB25} augments dense retrieval with Maximum Marginal Relevance (MMR)~\citep{carbonell1998use} to balance topical similarity against inter-example diversity, showing that small gains in diversity yield consistent improvements across prompt sizes. SuperICL~\citep{XuXWLZM24} couples a commercial LLM with a locally trained small language model, injecting the small model's predictions and confidence scores into the prompt so that the large model can directly reason over domain-specific signals. GenICL~\citep{zhang2025learning} departs from retrieval-based surrogate objectives, treating example selection as generative preference learning based on LLM feedback and training a reranker to identify examples that the downstream model itself judges useful. These approaches collectively show that ICL performance can be shaped substantially through example design rather than parameter updates. However, they primarily optimize example selection in generic NLP settings, without explicitly addressing the structured semantic alignment and open-set recognition required in safety-critical ATC readback monitoring. These limitations motivate a more task-oriented design of retrieval and reasoning for real environments.

\section{Problem Definition}
\label{sec:prob}
This section formalizes readback anomaly detection as an open-set classification problem under the ICL paradigm. The task requires the model to jointly recognize predefined readback classes and detect unknown communication patterns that are absent during training, while relying on retrieved examples for inference without parameter updates.

\begin{figure}[!htbp]
    \centering
    \includegraphics[width=\linewidth]{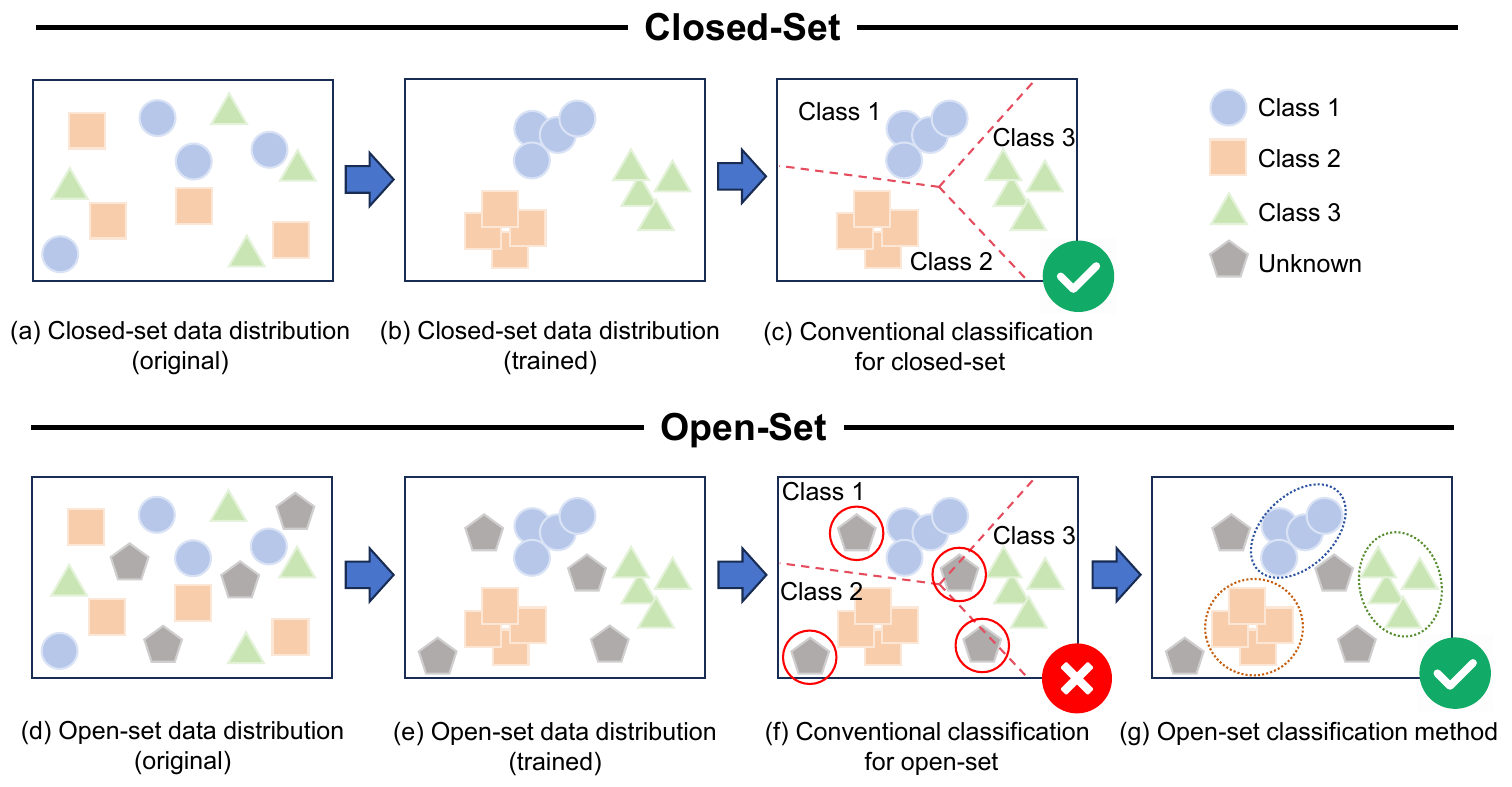}
    \caption{Comparison between closed-set and open-set classification.}
    \label{fig:Task schematic}
\end{figure}

\textbf{Open-Set Readback Classification.}
Let the training set be
\(
\mathcal{D}_{\mathrm{train}}
=
\{(x_i^{(1)}, x_i^{(2)}, y_i)\}_{i=1}^{I},
\)
where \(x_i^{(1)}\) and \(x_i^{(2)}\) denote the first (preceding) and second (subsequent) utterances in the \(i\)-th ATCo--pilot dialogue exchange, respectively, and \(y_i \in \{0,1\}^{K}\) is the corresponding one-hot label vector over \(K\) well-defined readback classes. In a classification problem, the learnable parameters \(\theta\) are optimized on \(\mathcal{D}_{\mathrm{train}}\) by minimizing the empirical classification risk,

\begin{equation}
\theta^*
=
\arg\min_{\theta}\,
\mathbb{E}_{(x_i^{(1)},x_i^{(2)},y_i)\in\mathcal{D}_{\mathrm{train}}}
\mathbb{I}\!\left(
\hat{y}_i \neq y_i
\right),
\end{equation}
where \(\hat{y}_i\) denotes the predicted one-hot label vector and \(\mathbb{I}(\cdot)\) is the indicator function.
Under the conventional closed-set assumption, both training and test samples are drawn from the same predefined label space. As illustrated in \hyperref[fig:Task schematic]{Fig. 2(a)}, \hyperref[fig:Task schematic]{2(b)}, and \hyperref[fig:Task schematic]{2(c)}, closed-set classification learns decision boundaries only among known classes and therefore is effective when all test samples belong to these well-defined classes.

However, real-world ATCo--pilot communications inevitably contain ambiguous or undefined utterance pairs that are not covered by the predefined readback classes. In this case, test samples can come from one of the \(K\) known classes or an additional unknown class that is absent during training. As shown in \hyperref[fig:Task schematic]{Fig. 2(d)}, \hyperref[fig:Task schematic]{2(e)}, and \hyperref[fig:Task schematic]{2(f)}, a conventional closed-set classifier tends to force unknown samples into known classes, leading to incorrect decisions near the open boundary. Therefore, readback monitoring requires the model to classify known readback types while detecting unknown communication patterns. Accordingly, the ground-truth label of the \(j\)-th test sample is defined over an open label space as \(y_j^{\mathrm{o}} \in \{0,1\}^{K+1}\), where the additional \((K+1)\)-th dimension denotes the unknown class, as illustrated in \hyperref[fig:Task schematic]{Fig. 2(g)}. The corresponding prediction is denoted by \(\hat{y}_j^{\mathrm{o}}\).

\begin{figure}[t!]
    \centering
    \includegraphics[width=\linewidth]{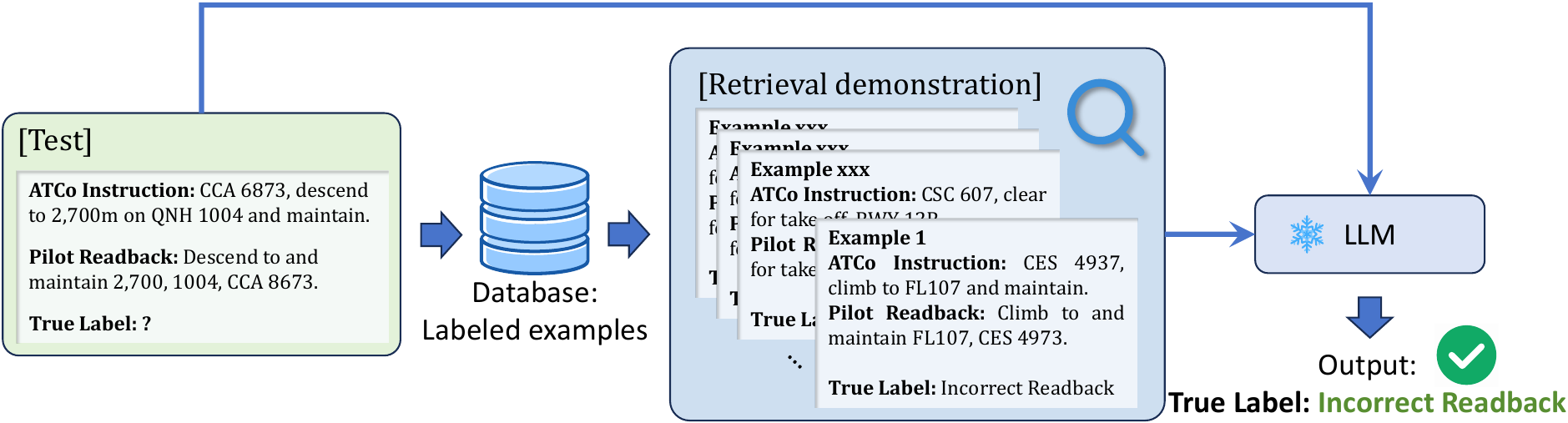}
    \caption{Illustration of an ICL task for ATC readback monitoring.}
    \label{fig:ICLtask}
\end{figure}

\textbf{In-Context Learning.} Without any additional training, ICL relies on retrieved examples in the support set to provide task-specific patterns for in-context label inference. Accordingly, an illustration of ICL in \hyperref[fig:ICLtask]{Fig. 3} shows that, the support set \(\mathcal{D}_{\mathrm{support}}\) is constructed by retrieving $N$ examples per class from \(\mathcal{D}_{\mathrm{train}}\) based on the test sample \(x_j = (x_j^{(1)}, x_j^{(2)})\),
\begin{equation}
    \mathcal{D}_{\mathrm{support}}(x_j) = 
    \{(x_s^{(1)}, x_s^{(2)}, y_s)\}_{s=1}^{N \times (K+1)}
    \subset \mathcal{D}_{\mathrm{train}},
\end{equation}
where \(\mathcal{D}_{\mathrm{support}}\) contains \(N\) examples per class, covering all \(K+1\) classes in the extended label space. Subsequently, a generative language model $\pi$ takes $\mathcal{D}_{\mathrm{support}}$ as context and predicts the classification label $\hat{y}^{\mathrm{o}}_j$ by
\begin{equation}
    \hat{y}^{\mathrm{o}}_j
    =
    \arg\max_{y_j^{\mathrm{o}}}
    \mathcal{P}_{\pi}\!\left(
    y_j^{\mathrm{o}}
    \mid
    \mathcal{D}_{\mathrm{support}},
    x_j^{(1)}, x_j^{(2)}
    \right),
\end{equation}
where \(\mathcal{P}_{\pi}\) denotes the conditional probability distribution over labels assigned by the model \(\pi\) given the test input and support set.

\section{Methodology}\label{sec:method}

\begin{figure}[!htbp]
    \centering
    \includegraphics[width=0.9\linewidth]{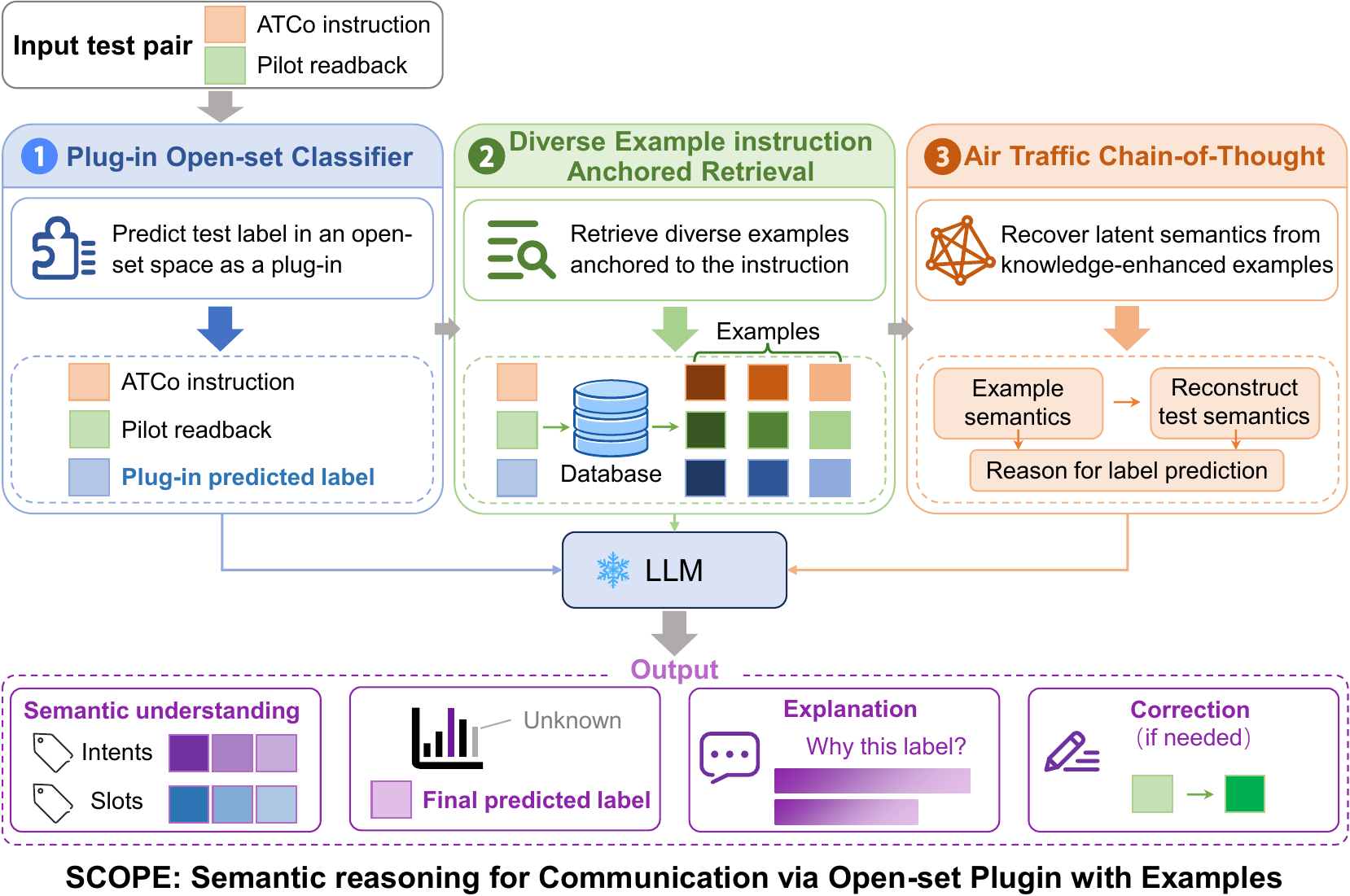}
    \caption{The architecture and key modules of the proposed framework SCOPE.}
    \label{fig:method framework}
\end{figure}

In this section, we present the proposed SCOPE framework for open-set ATC readback monitoring, which produces predicted readback class labels, semantic explanations, and correction candidates. As illustrated in \hyperref[fig:method framework]{Fig. 4}, SCOPE consists of three key modules. First, a Plug-in Open-set Classifier (POC) provides stable prior predictions and enables reliable detection of unknown communication patterns, thereby coupling the LLM's semantic reasoning ability with the classification strength of a lightweight model. Second, a Diverse Example instruction Anchored Retrieval (DEAR) module selects scenario-relevant and class-discriminative examples to support ICL. Third, the Air Traffic Chain-of-Thought (ATCoT) module performs structured semantic reasoning to recover latent intent and slot information, enabling decisions that better reflect ATC operational logic. Building upon these processes, the system further generates interpretable explanations and produces standard corrections through rule-guided semantic reordering.

\subsection{Plug-in Open-set Classifier}

\begin{figure}[!htbp]
    \centering
    \includegraphics[width=0.8\linewidth]{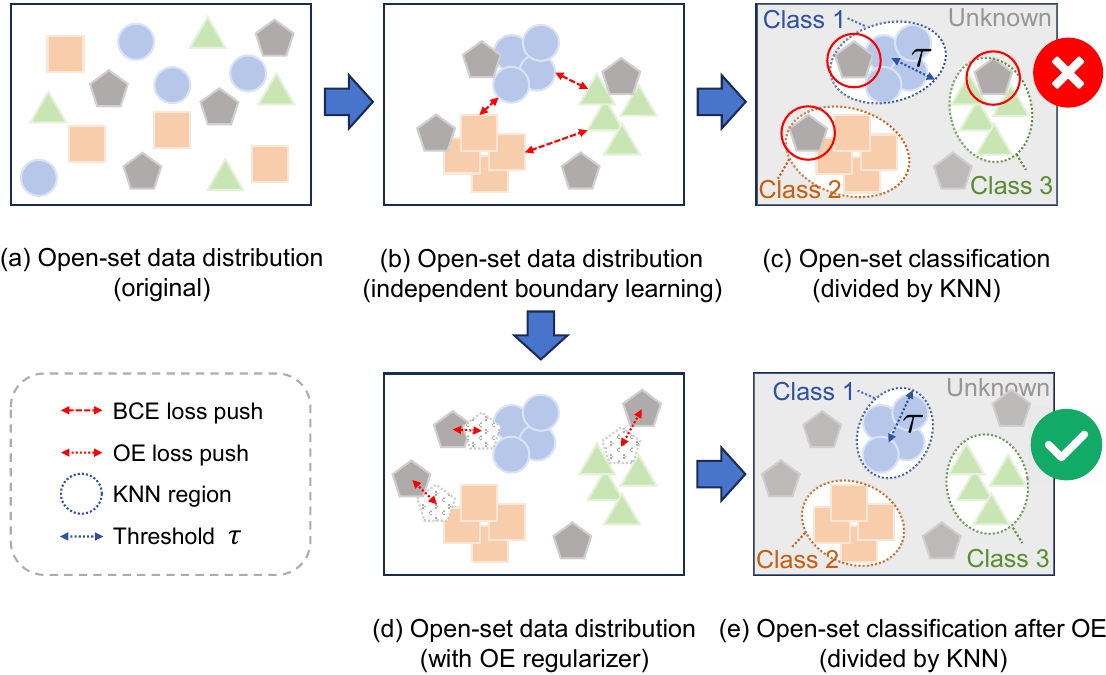}
    \caption{Illustration of POC. Independent boundary learning first forms separable regions for known readback classes in the feature space, while OE regularization further pushes unknown samples away from known-class regions. KNN with threshold \(\tau\) then detects communication pairs that fall outside the known readback space.}
    \label{fig:module 1}
\end{figure}

We propose the Plug-in Open-set Classifier (POC), a lightweight model for open-set readback recognition, as shown in \hyperref[fig:module 1]{Fig. 5}. It assigns an independent probability to each known class rather than forcing all classes to compete under a shared softmax. We employ an Outlier Exposure (OE) regularization component to reduce overconfident assignment of unknown inputs to known readback classes~\citep{DBLP:conf/iclr/HendrycksMD19}. This is followed by KNN-based geometric detection at test time to identify ATCo--pilot communication pairs that fall outside the known readback space~\citep{cover1967nearest}.

Given an utterance pair \(x_i=(x_i^{(1)},x_i^{(2)})\), a Transformer-based encoder maps it into a semantic feature,
\begin{equation}
h(x_i)=f(x_i^{(1)},x_i^{(2)};\theta),
\label{eq:hidden size}
\end{equation}
where \(f(\cdot;\theta)\) denotes a DeBERTa encoder~\citep{he2021deberta}, which improves contextual representation by disentangling content and positional information in self-attention. The resulting feature \(h(x_i) \in \mathbb{R}^{d}\) is the joint representation of the utterance pair, where \(d\) is the hidden size.

To model the known readback space and learn class boundaries, we adopt a sigmoid layer on top of \(h(x_i)\). This layer outputs an independent probability for each class, allowing relatively separate acceptance regions to emerge in the shared semantic space. Specifically, it is parameterized by \(W_{\mathrm{doc}}\in\mathbb{R}^{K\times d}\) and \(b_{\mathrm{doc}}\in\mathbb{R}^{K}\), which map \(h(x_i)\) to a \(K\)-dimensional per-class probability vector \(p_i\in(0,1)^K\), followed by independent sigmoid activations,
\begin{equation}
p_i=\sigma \bigl(W_{\mathrm{doc}}h(x_i)+b_{\mathrm{doc}}\bigr),
\end{equation}
where \(\sigma(\cdot)\) denotes the element-wise sigmoid function. This formulation allows the model to evaluate each readback class independently.

POC is trained with a Binary Cross-Entropy (BCE) loss over the $I$ labeled training samples,
\begin{equation}
\mathcal{L}_{\mathrm{BCE}}
=
-\frac{1}{IK}\sum_{i=1}^{I}\sum_{k=1}^{K}
\left[
y_i^k\log p_i^k + (1-y_i^k)\log(1-p_i^k)
\right],
\end{equation}
where \(y_i^k \in \{0,1\}\) is the binary label indicating whether sample $x_i$ belongs to class $k$, and \(p_i^k\) denotes the predicted probability that this sample belongs to the \(k\)-th class.

However, since POC is trained only on known classes, it may assign spuriously high known-class probabilities to unknown inputs. We therefore introduce an OE regularizer with an unknown set \(\mathcal{D}_{\mathrm{UK}}=\{x_u\}_{u=1}^{U}\), consisting of non-readback samples. These samples are used to make the known-class boundaries more conservative, encouraging the model to stay uncertain on unknown inputs instead of forcing them into a specific known class. Concretely, for each \(x_u\), let \(p_u \in (0,1)^K\) denote its sigmoid outputs over the \(K\) known classes. We minimize
\begin{equation}
\mathcal{L}_{\mathrm{OE}}
=
-\frac{1}{U K}
\sum_{u=1}^{U}\sum_{k=1}^{K}
\left[
t\log p_u^k + (1-t)\log(1-p_u^k)
\right],
\label{eq:oe loss}
\end{equation}
where \(t \in [0,1]\) is the exposure target. Setting \(t=0.5\) gives equal weights to both log terms. The loss is minimized when \(p_u^k=0.5\) for all \(k\), which drives the model toward maximum uncertainty with respect to all known classes. It is worth noting that \(\mathcal{D}_{\mathrm{UK}}\) is not used to train an additional \((K+1)\)-th classifier for the unknown class. Instead, it serves only as exposure data for boundary regularization. The model is therefore not encouraged to learn a fixed unknown-class prototype, but to avoid confident assignment of non-readback samples to any known readback class. This distinction is important because unknown communication patterns are open-ended and cannot be represented by a single supervised class.

Taking into account both BCE and OE losses, the final training objective is defined as their weighted sum,
\begin{equation}
\mathcal{L}
=
\mathcal{L}_{\mathrm{BCE}} + \lambda\,\mathcal{L}_{\mathrm{OE}},
\end{equation}
where $\lambda$ is a hyperparameter that balances the contributions of the two loss terms.

Distinguishing known from unknown samples fundamentally depends on the geometry of the semantic space. Although OE suppresses overconfident predictions on unknown samples, it does not explicitly regularize the learned feature space. Therefore, we incorporate a KNN geometric detection module. Specifically, a test sample is classified as known if it lies sufficiently close to the known feature bank; otherwise, it is assigned to the unknown class.

The cosine similarity between two feature representations \(h(x_j)\) and \(h(x_i)\) is defined as
\begin{equation}
\mathrm{SIM}\bigl(h(x_j), h(x_i)\bigr)
=
\frac{h(x_j)^\top h(x_i)}
{\|h(x_j)\|_2\,\|h(x_i)\|_2},
\label{eq:sim}
\end{equation}
where \(h(x)\in\mathbb{R}^{d}\) denotes the feature extracted by the trained encoder \(f(\cdot;\theta)\) in \hyperref[eq:hidden size]{Eq. (4)}, \((\cdot)^\top\) denotes transposition, and \(\|\cdot\|_2\) denotes the Euclidean norm.

The reference feature bank of known classes \(\mathcal{H}_{\mathrm{train}}=\{h(x_i)\}_{i=1}^{I}\) is constructed from the training set, and the KNN score of a test sample \(x_j\) is given by
\begin{equation}
s(x_j)
=
\kappa\text{-}\min_{h(x_i)\in\mathcal{H}_{\mathrm{train}}}
\left(
1-\mathrm{SIM}\bigl(h(x_j), h(x_i)\bigr)
\right),
\end{equation}
where \(1-\mathrm{SIM}\bigl(h(x_j), h(x_i)\bigr)\) represents the cosine distance between the test feature and a training feature. \(\kappa\text{-}\min\) returns the \(\kappa\)-th smallest value among all candidate distances. A small \(s(x_j)\) indicates proximity to the known readback region, whereas a large value suggests that the sample lies outside the valid readback domain and should be assigned to the unknown class.

POC performs open-set prediction by combining classification with distance detection,
\begin{equation}
\hat{y}_j^{\mathrm{o}}=
\begin{cases}
K+1, & s^{\mathrm{knn}}(x_j)>\tau,\\
\displaystyle \arg\max_{k\in\{1,\dots,K\}} p_j^k, & s^{\mathrm{knn}}(x_j)\le\tau,
\end{cases}
\end{equation}
where \(\tau\) is an iteratively calibrated decision threshold for KNN-based unknown detection. In this way, POC produces unified open-set predictions for ATCo--pilot communication pairs, including both known readback classes and the unknown class.

As a widely accepted criterion for threshold selection, Youden's \(J\) statistic~\citep{youden1950index} is used to determine the KNN threshold \(\tau\),
\begin{equation}
J(\tau)
=
\frac{1}{|\mathcal{D}^{\mathrm{UK}}|}\!
\sum_{x\in\mathcal{D}^{\mathrm{UK}}}\!
\mathbb{I}_{\tau}(x)
-
\frac{1}{|\mathcal{D}^{\mathrm{K}}|}\!
\sum_{x\in\mathcal{D}^{\mathrm{K}}}\!
\mathbb{I}_{\tau}(x),
\end{equation}
where \(\mathbb{I}_{\tau}(x)\) is an indicator function that equals \(1\) if \(s^{\mathrm{knn}}(x)>\tau\), and \(0\) otherwise. The first term measures the proportion of unknown samples \(\mathcal{D}^{\mathrm{UK}}\) that are correctly detected as unknown, whereas the second term measures the proportion of known samples \(\mathcal{D}^{\mathrm{K}}\) that are incorrectly assigned to the unknown class. Accordingly, \(J(\tau)\) characterizes the trade-off between unknown detection and the preservation of valid readback samples under threshold \(\tau\). Finally, the optimal threshold is given by \(\tau^*=\arg\max_{\tau} J(\tau)\).

\begin{figure}[!htbp]
    \centering
\includegraphics[width=0.9\linewidth]{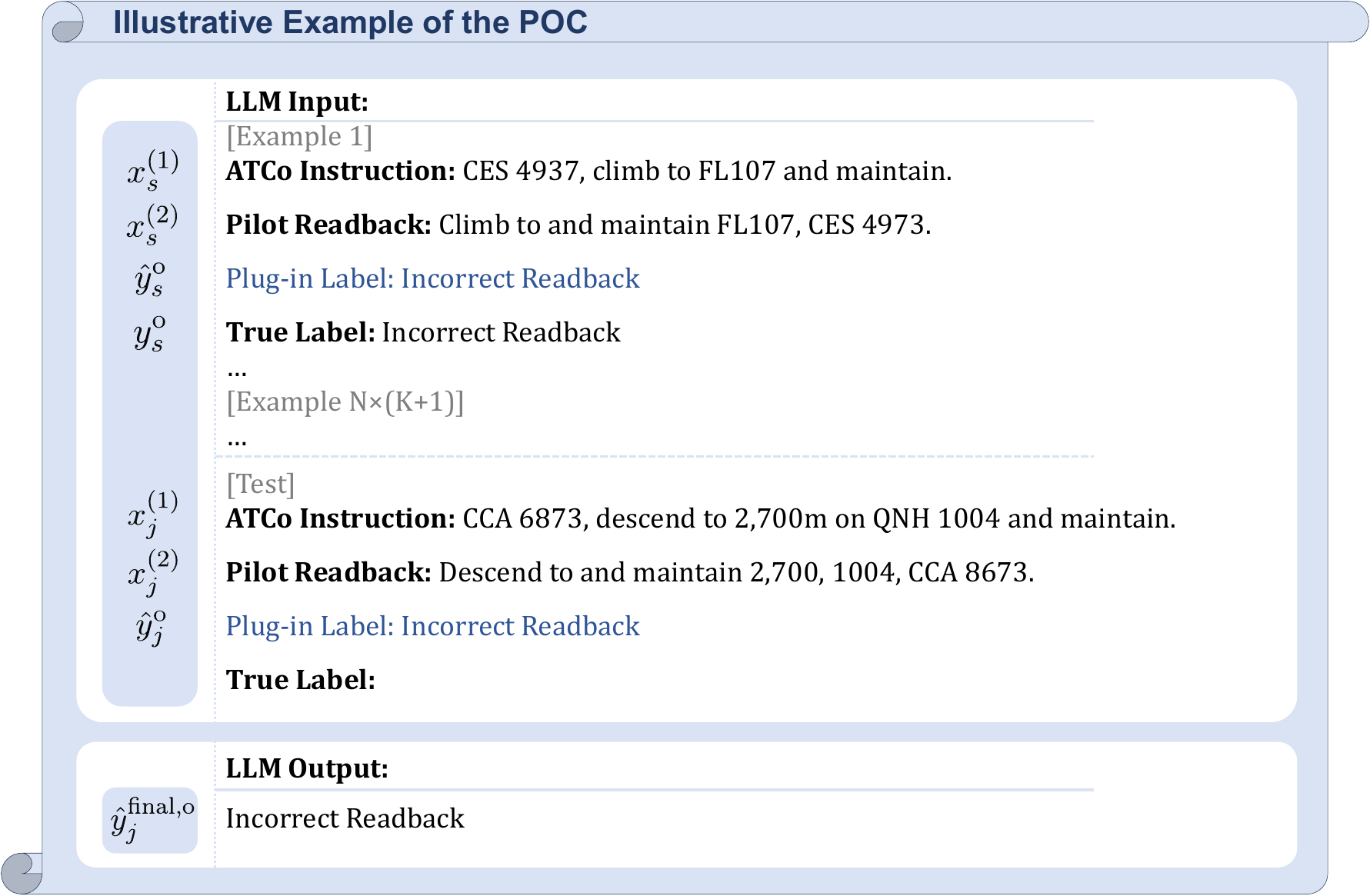}
    \caption{An example of POC during LLM inference.}
    \label{fig:module1example}
\end{figure}

The trained POC is further integrated into the subsequent LLM inference stage. Specifically, as illustrated in \hyperref[fig:module1example]{Fig. 6}, its predicted label \(\hat{y}_j^{\mathrm{o}}\) for the test sample \(x_j\) is injected into LLM as plug-in label in the prompt. In doing so, the LLM performs semantic reasoning under an explicit prior over the possible readback state, thereby gaining an initial ability to discriminate between known readback classes and unknown communication patterns without introducing model complexity.

\subsection{Diverse Example Instruction Anchored Retrieval}
\begin{figure}[htbp]
    \centering
    \includegraphics[width=\linewidth]{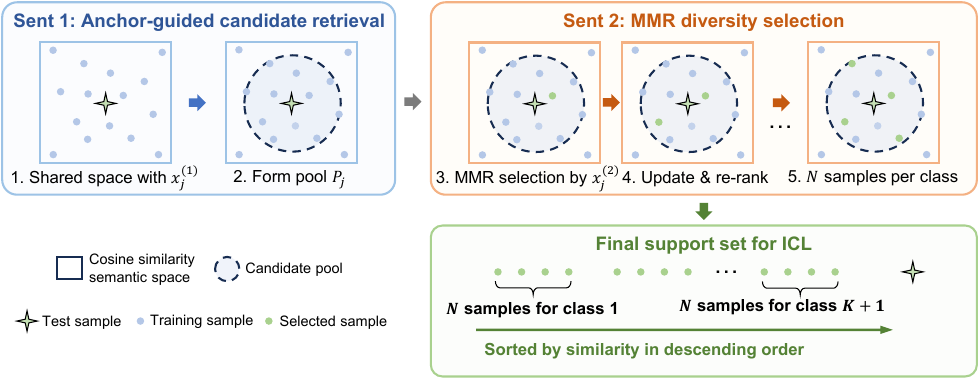}
    \caption{Illustration of DEAR. The ATCo instruction \(x_j^{(1)}\) is first used as an anchor to retrieve a scenario-relevant candidate pool \(P_j\) from the training set. Within each label class, MMR then selects examples according to the pilot readback \(x_j^{(2)}\), balancing relevance to the test sample and diversity among selected examples. The final support set contains \(N\) examples per class and is sorted by anchor similarity before being used for ICL.}
    \label{fig:module 2}
\end{figure}

In ICL, retrieved support samples are provided in the input context. The approach is effective because the test sample interacts with these examples through self-attention in an LLM composed of stacked Transformer layers, allowing the model to form a context-conditioned representation~\citep{yang2024context}.
Therefore, the selection and ordering of examples substantially affect ICL performance, and poorly chosen examples can lead to degenerate or inconsistent predictions. As illustrated in \hyperref[fig:module 2]{Fig. 7}, we design the Diverse Example instruction Anchored Retrieval (DEAR) method. It exploits the inherent asymmetry of ATCo--pilot communications, while explicitly balancing anchor relevance and intra-class diversity to provide richer example references.

Within the known readback classes, the preceding utterance is typically an ATCo instruction, which conveys the operational intent to be reflected in the subsequent pilot readback. The preceding utterance \(x_j^{(1)}\) is therefore treated as the test anchor, also referred to as the ``gold sentence''. Both the test anchor and the training preceding utterances \(\{x_i^{(1)}\}_{i=1}^{I} \subset \mathcal{D}_{\mathrm{train}}\) are encoded into a shared high-dimensional retrieval space, yielding representations \(\mathfrak{h}(x_j^{(1)}) \in \mathbb{R}^{d_1}\) and \(\mathfrak{h}(x_i^{(1)}) \in \mathbb{R}^{d_1}\), respectively. The scenario relevance between the test anchor and the \(i\)-th training preceding utterance is then quantified by \(\mathrm{SIM}\bigl(\mathfrak{h}(x_j^{(1)}), \mathfrak{h}(x_i^{(1)})\bigr)\) in \hyperref[eq:sim]{Eq. (9)}. A higher similarity score indicates a closer operational context and thus stronger scenario relevance.

For the test anchor $x_j^{(1)}$, all training preceding utterances $\{x_i^{(1)}\}_{i=1}^{I}$ are ranked in descending order according to scenario relevance. The top $M$ retrieved utterance pairs are retained to form the candidate pool,
\begin{equation}
P_j
=
\left\{
\left(x_i^{(1)}, x_i^{(2)},\hat{y}_i, y_i\right)
\;\middle|\;
\operatorname{rank}_{(x_j)}(i)\le M
\right\},
\end{equation}
where \(M=\rho N\), \(\rho\) denotes a preset pool ratio, and \(\operatorname{rank}_{(x_j)}(i)\) denotes the rank of the \(i\)-th training sample according to its scenario relevance to the \(j\)-th test sample $x_j$. This anchoring step restricts the subsequent search to a semantically coherent candidate pool and reduces the computational burden of the downstream greedy selection procedure. Consequently, low-relevance examples are filtered out before redundancy-aware selection is performed.

The pilot readback carries the fine-grained response pattern that is most directly related to class discrimination. Therefore, the final examples are selected from $P_j$ based on the subsequent utterance. Maximal Marginal Relevance (MMR) is adopted for greedy iterative selection~\citep{KapuriyaKGB25}. It favors candidates that are relevant to the test sample while discouraging redundancy with examples that have already been selected. Let $P'_j \subseteq P_j$ denote the set of examples already selected via the MMR criterion for the $j$-th test sample, starting with \(P'_j = \emptyset\). For each candidate $p \in P_j \setminus P'_j$, the MMR score is defined as
\begin{equation}
\operatorname{MMR}(p \mid P'_j)
=
\alpha \, \mathrm{SIM}\bigl(\mathfrak{h}(x_j^{(2)}), \mathfrak{h}(x_p^{(2)})\bigr)
-
(1-\alpha)\max_{p' \in P'_j}
\mathrm{SIM}\bigl(\mathfrak{h}(x_p^{(2)}), \mathfrak{h}(x_{p'}^{(2)})\bigr),
\end{equation}
where $\alpha \in [0,1]$ controls the tradeoff between relevance to the test readback and redundancy reduction among the selected examples. The MMR score is used as the selection criterion in an iterative greedy procedure. Specifically, the candidate with the highest score is selected at each step, appended to \(P'_j\), and the remaining candidates are re-ranked accordingly. This process continues until the desired number of examples is collected.

Once the MMR selection is completed, the retrieved examples from the known and unknown classes are assembled into the final support set for ICL. Formally,
\begin{equation}
\mathcal{D}_{\mathrm{support}}(x_j)
=
\operatorname{Sort}_{\downarrow}^{(1)}
\left(
\bigcup_{k=1}^{K+1}
\left\{
(x_{s,k}^{(1)},x_{s,k}^{(2)},\hat y^\mathrm{o}_{s,k},y^\mathrm{o}_{s,k})
\right\}_{s=1}^{N}
\right),
\end{equation}
where \((x_{s,k}^{(1)},x_{s,k}^{(2)},\hat y^\mathrm{o}_{s,k},y^\mathrm{o}_{s,k})\) denotes the \(s\)-th selected example for the \(k\)-th class with respect to the current test sample \(x_j\). The operator \(\operatorname{Sort}_{\downarrow}^{(1)}(\cdot)\) sorts all selected examples in descending order according to the scenario relevance \(\mathrm{SIM}\bigl(\mathfrak{h}(x_j^{(1)}), \mathfrak{h}(x_s^{(1)})\bigr)\).

\subsection{Air Traffic Chain-of-Thought}

Inspired by chain-of-thought and tree-of-thought reasoning~\citep{DBLP:conf/nips/YaoYZS00N23, DBLP:conf/nips/Wei0SBIXCLZ22}, we propose Air Traffic Chain-of-Thought (ATCoT) to refine open-set readback prediction through semantic reasoning. Instead of directly asking the LLM to predict the final label from the utterance pair alone, ATCoT provides semantically annotated examples, and then guides the model to reassess the test sample by eliciting ATC semantic knowledge.

\begin{figure}[!h]
    \centering
    \includegraphics[width=\linewidth]{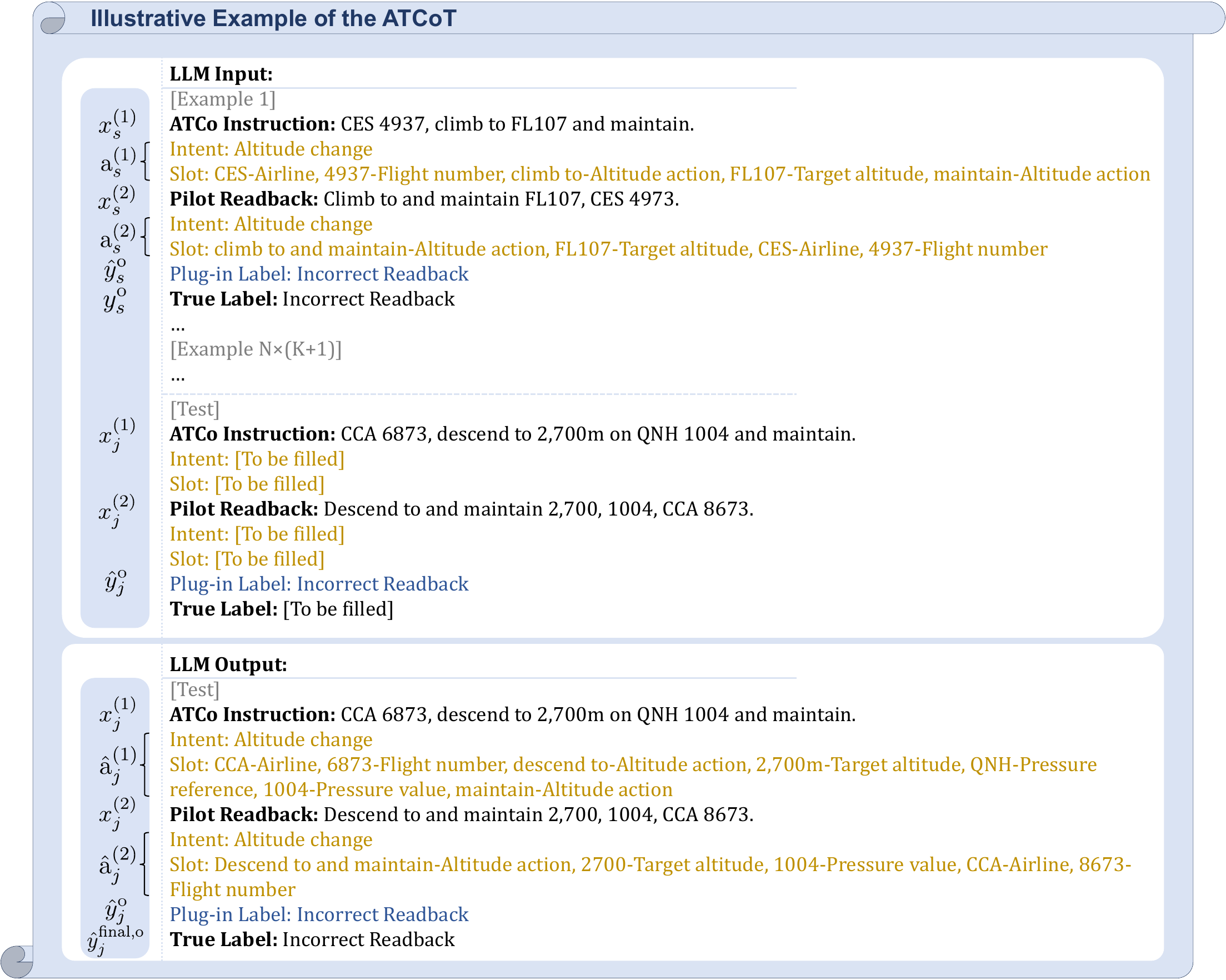}
    \caption{An example of the ATCoT during LLM inference.}
    \label{fig:module3example}
\end{figure}

\hyperref[fig:module3example]{Fig. 8} illustrates the resulting prompt layout and reasoning flow through an example. For each retrieved example, we attach a semantic annotation \(\mathrm{a}_s = \bigl(\mathrm{a}^{(1)}_s, \mathrm{a}^{(2)}_s\bigr)\), where \(\mathrm{a}^{(1)}_s = \bigl(\mathrm{i}^{(1)}_s, \mathrm{s}^{(1)}_s\bigr)\) and \(\mathrm{a}^{(2)}_s = \bigl(\mathrm{i}^{(2)}_s, \mathrm{s}^{(2)}_s\bigr)\) denote the intent and slot set associated with the ATCo instruction \(x_s^{(1)}\) and the pilot readback \(x_s^{(2)}\), respectively. The intent specifies the operational function of an utterance, while the slots encode its critical operational parameters and flight-related entities. Accordingly, the semantically enriched support set is written as
\begin{equation}
\widetilde{\mathcal{D}}_{\mathrm{support}}(x_j)
=
\left\{
\bigl(
x_s^{(1)},\,
\mathrm{a}^{(1)}_s,\,
x_s^{(2)},\,
\mathrm{a}^{(2)}_s,\,
\hat{y}^{\mathrm{o}}_s,\,
y^{\mathrm{o}}_s
\bigr)
\right\}_{s=1}^{|\mathcal{D}_{\mathrm{support}}(x_j)|},
\end{equation}
where \(\hat{y}^{\mathrm{o}}_s\) and \(y^{\mathrm{o}}_s\) denote the predicted and ground-truth open-set labels of the \(s\)-th example, respectively.

At inference time, unlike the examples, the test sample is not supplied with any intent or slot semantic annotation in advance. Therefore, its semantic structure must be inferred by the LLM during reasoning. 
We treat the test semantic annotation \(\mathrm{a}_j = \bigl(\mathrm{a}_j^{(1)}, \mathrm{a}_j^{(2)}\bigr)\) as a unified latent variable to be inferred by the LLM. Given the semantically enriched support set, the test utterance pair, and the prior label predicted, ATCoT models the joint distribution of the latent semantics and the final label as
\begin{equation}
\mathcal{P}_{\pi}\!\left(
\mathrm{a}_j, y_j^{\mathrm{o}}
\mid
\widetilde{\mathcal{D}}_{\mathrm{support}}(x_j),
x_j^{(1)}, x_j^{(2)},
\hat{y}_j^{\mathrm{o}}
\right).
\end{equation}

As a result, the model is encouraged to recover the structured semantics underlying the utterance pair before committing to the final label, thereby enhancing semantic reasoning and improving prediction accuracy.

The corresponding inference seeks the most plausible semantic reconstruction and label prediction,
\begin{equation}
\left(\hat{\mathrm{a}}^{(2)}_j, \hat{y}^{\text{final},\mathrm{o}}_j\right)
=
\arg\max_{\mathrm{a}_j,\,{y}^{\mathrm{o}}_j}
\mathcal{P}_{\pi}\!\left(
\mathrm{a}_j, {y}^{\mathrm{o}}_j
\mid
\widetilde{\mathcal{D}}_{\mathrm{support}}(x_j),
x_j^{(1)}, x_j^{(2)},
\hat{y}^{\mathrm{o}}_j
\right).
\end{equation}

By the chain rule, this joint distribution can be written as
\begin{equation}
\mathcal{P}_{\pi}\!\left(
\mathrm{a}_j, {y}^{\mathrm{o}}_j
\mid \cdot
\right)
=
\mathcal{P}_{\pi}\!\left(
\mathrm{a}_j
\mid \cdot
\right)
\,
\mathcal{P}_{\pi}\!\left(
{y}^{\mathrm{o}}_j
\mid
\mathrm{a}_j, \cdot
\right),
\end{equation}
where \(\cdot\) abbreviates the context. The first term measures how well the recovered semantics are supported by the input context, and the second term scores the final label under the reconstructed semantics.

The final label distribution is obtained by marginalizing over the latent semantic variable,
\begin{equation}
\mathcal{P}_{\pi}\!\left(
{y}^{\mathrm{o}}_j
\mid
\widetilde{\mathcal{D}}_{\mathrm{support}}(x_j),
x_j^{(1)}, x_j^{(2)},
\hat{y}^{\mathrm{o}}_j
\right)
=
\sum_{\mathrm{a}_j}
\mathcal{P}_{\pi}\!\left(
\mathrm{a}_j, {y}^{\mathrm{o}}_j
\mid
\widetilde{\mathcal{D}}_{\mathrm{support}}(x_j),
x_j^{(1)}, x_j^{(2)},
\hat{y}^{\mathrm{o}}_j
\right).
\end{equation}

Accordingly, the final prediction is
\begin{equation}
\hat{y}^{\text{final},\mathrm{o}}_j
=
\arg\max_{y \in \{1,\dots,K+1\}}
\mathcal{P}_{\pi}\!\left(
y
\mid
\widetilde{\mathcal{D}}_{\mathrm{support}}(x_j),
x_j^{(1)}, x_j^{(2)},
\hat{y}^{\mathrm{o}}_j
\right).
\end{equation}

\subsection{Explanation and Correction}

Building upon anomaly detection in readback monitoring, we include explanation and correction outputs to enhance the operational practicality of the proposed system. In real-world ATC support, a detector that only outputs a class label could be insufficient. It is essential to indicate the semantic basis of each decision and, when an anomaly is detected, provide a standard correction candidate. 
To this end, the framework extends readback detection into structured explanation and correction, forming a cascaded pipeline for ATC readback monitoring. The explanation is generated by the same LLM used for anomaly detection, ensuring consistency in reasoning, while the correction is derived from slots extracted by ATCoT and refined according to ATC phraseology rules.

\subsubsection{Explanation by LLM} 
To improve the usability of the anomaly detection output, we adopt a structured explanation schema. Guided by the semantically enhanced support set \(\widetilde{\mathcal{D}}_{\mathrm{support}}(x_j)\) and the inferred semantic annotation \(\hat{\mathrm{a}}_j\), the explanation is organized around key operational dimensions, including intent alignment and slot consistency. This design enables the system to make the rationale behind its predictions explicit using semantic evidence, rather than relying on a black-box decision alone. To this end, the decision process becomes easier to inspect and the overall system gains greater transparency. Based on this reconstructed semantic representation, the system generates an explanation \(e_j\) for the predicted label,
\begin{equation}
e_j
=
\arg\max_{e}
\mathcal{P}_{\pi}\!\bigl(
e
\mid
\widetilde{\mathcal{D}}_{\mathrm{support}}(x_j),\,
x_j^{(1)},\, x_j^{(2)},\,
\hat{\mathrm{a}}_j,\,
\hat{y}_j^{\mathrm{o}},\,
\hat{y}^{\text{final},\mathrm{o}}_j
\bigr),
\end{equation}
where \(e_j\) summarizes the semantic evidence underlying the decision. Here, \(\pi\) denotes the same LLM used for anomaly detection, ensuring that explanation generation is consistent with the decision process and grounded in the same reasoning mechanism.

\subsubsection{Correction by Semantic Reordering}
When the final predicted label \(\hat{y}^{\text{final},\mathrm{o}}_j\) indicates an anomalous readback, the system generates a corrected readback candidate. Rather than relying on free form generation, we employ a deterministic correction strategy grounded in the inferred semantics, which better aligns with the standardized nature of ATC phraseology~\citep{icao_doc9432}.
Specifically, when an anomalous readback is detected, the framework generates a corrected readback candidate through three steps. First, the slots inferred by the structured reasoning module are aligned with those in the ATCo instruction. Second, the airline designator and flight number are extracted and positioned at the end of the readback in accordance with phraseology conventions. Finally, the corrected readback is generated by composing the reordered semantic slots into a well-formed sentence.

Formally, the system uses the inferred semantic structure \(\hat{\mathrm{a}}_j^{(1)}\) to locate the callsign span in the ATCo instruction \(x_j^{(1)}\), removes it from its original position, and appends it to the end of the utterance to form a canonical corrected readback,
\begin{equation}
\hat{x}_j^{(2)}
=
\mathcal{C}\bigl(
x_j^{(1)},\,
\hat{\mathrm{a}}_j^{(1)}
\bigr),
\end{equation}
where \(\mathcal{C}(\cdot)\) denotes a deterministic reordering operator. It extracts the callsign from \(x_j^{(1)}\) according to \(\hat{\mathrm{a}}_j^{(1)}\), keeps the remaining instruction content unchanged, and places the callsign at the sentence end. By doing so, correction is implemented as semantic reconstruction rather than unconstrained generation, which reduces hallucination risk and improves interpretability.

\section{Experiments}
\label{sec:experiment}
This section systematically evaluates the proposed SCOPE framework under safety-critical ATC scenarios. After introducing the datasets, experimental settings, baselines, and evaluation metrics, we assess the framework from both model and system perspectives, including overall readback monitoring performance, the underlying mechanisms of each module, and deployment feasibility in real operational environments.

\subsection{Experimental Settings}

\subsubsection{Dataset}

The dataset used in this study is a subset of the Air Traffic Spoken Instruction Understanding (ATSIU) dataset, which contains real ATCo--pilot communications collected from multiple airports and airspace sectors across China~\citep{zhang2025atsiu}. The raw voice recordings were transcribed into text by trained ATCos. From the transcripts, we extracted two-turn utterance pairs, each consisting of one preceding utterance and its immediately following response.
To characterize readback deviations in a manner consistent with human factor analysis, we adopt a taxonomy that distinguishes among errors, lapses, and slips~\citep{icao_doc9683}. Based on this taxonomy, the present study considers four in-domain readback classes: \emph{Correct Readback}, \emph{Incorrect Readback}, \emph{Incomplete Readback}, and \emph{Non-standard Readback}. However, anomalous readbacks are difficult to collect at sufficient scale from real operations. To address this limitation, we construct the latter three classes from authentic correct readbacks by introducing deviations from standard radiotelephony~\citep{icao_doc9432}. Specifically, \emph{Incorrect Readback} samples are created by modifying safety-critical operational elements, \emph{Incomplete Readback} samples are created by removing one or more required elements, and \emph{Non-standard Readback} samples are created by introducing operationally realistic non-standard phraseology or altered word order that remains interpretable but departs from standard communications.
In addition, real communications contain many utterance pairs that do not belong to the readback process, such as clarification requests, routine inquiries, acknowledgments, and other non-readback exchanges. These cases are grouped into an \emph{Unknown} class. This formulation better reflects operational reality and naturally casts the task as open-set readback monitoring. Based on this design, we name the resulting dataset ATCo--Pilot Communication Pairs (APCP). \hyperref[tab:class_distribution]{Table 2} shows the class distributions of various subsets.

\begin{table}[htbp]
\centering
\caption{Class distributions of the APCP training, test, and calibration sets. The calibration set corresponds to \(\mathcal{D}_{\mathrm{UK}}\), which is used for OE regularization and Youden's \(J\) threshold calibration.}
\label{tab:class_distribution}
\begin{tabular}{lccc}
\toprule
Class & Train set & Test set & Calibration set \\
\midrule
Correct  & 1,299 (52.9\%) & 547 (48.0\%) & 0 (0.0\%) \\
Incorrect    & 622 (25.3\%)   & 268 (23.5\%) & 0 (0.0\%) \\
Incomplete  & 376 (15.3\%)   & 167 (14.6\%) & 0 (0.0\%) \\
Non-standard & 159 (6.5\%)    & 72 (6.3\%)   & 0 (0.0\%) \\
Unknown  & 0 (0.0\%)      & 86 (7.5\%)   & 204 (100.0\%) \\
\midrule
Total    & 2,456 & 1,140 & 204 \\
\bottomrule
\end{tabular}
\end{table}

\begin{figure}[t!]
    \centering
    \includegraphics[width=0.75\linewidth]{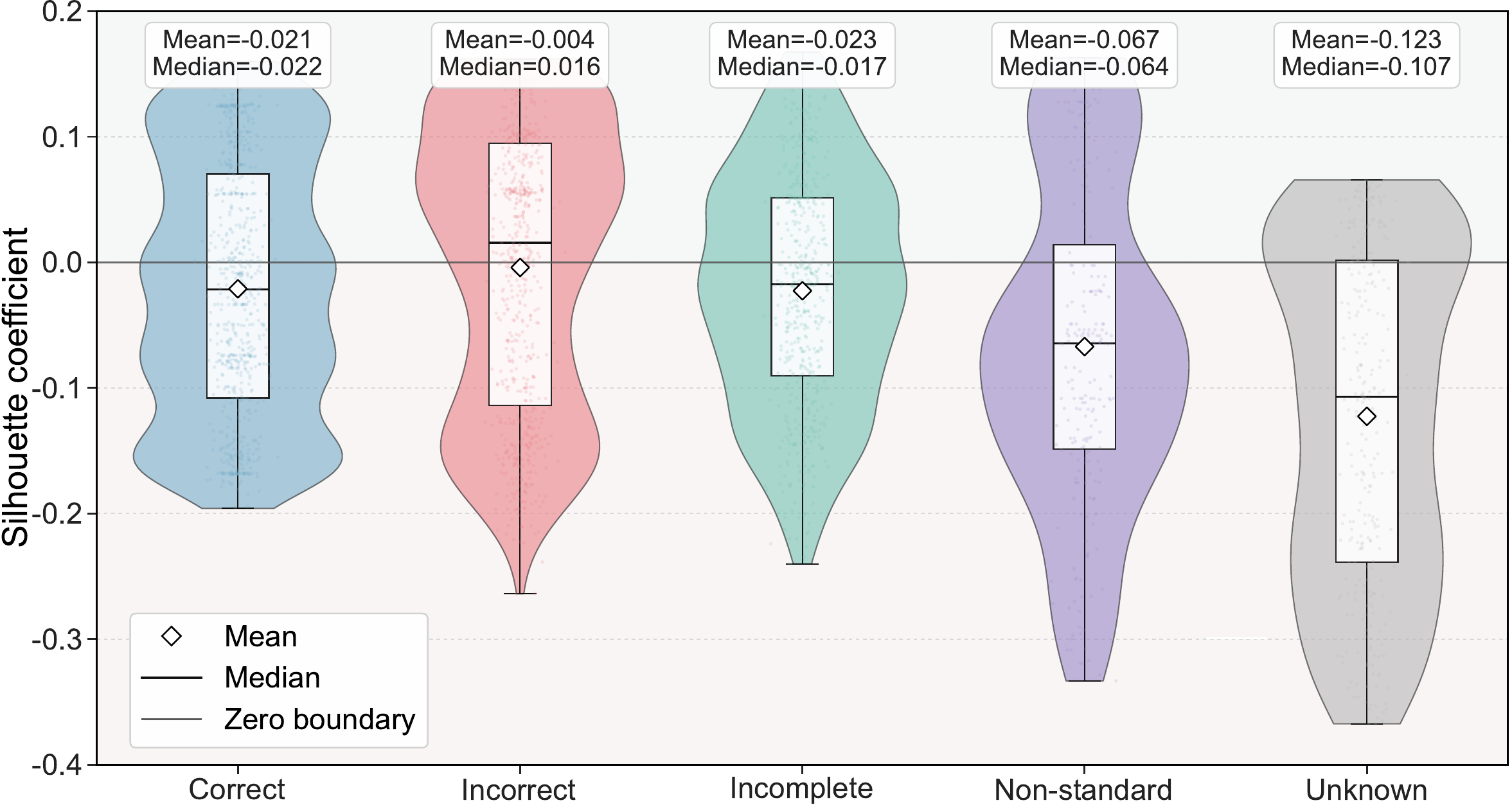}
    \caption{Violin plot of original silhouette coefficients. Larger silhouette coefficients indicate better intra-class compactness and inter-class separability in the representation space.}
    \label{fig:silhouette}
\end{figure}

\begin{figure}[t!]
    \centering
    \includegraphics[width=0.75\linewidth]{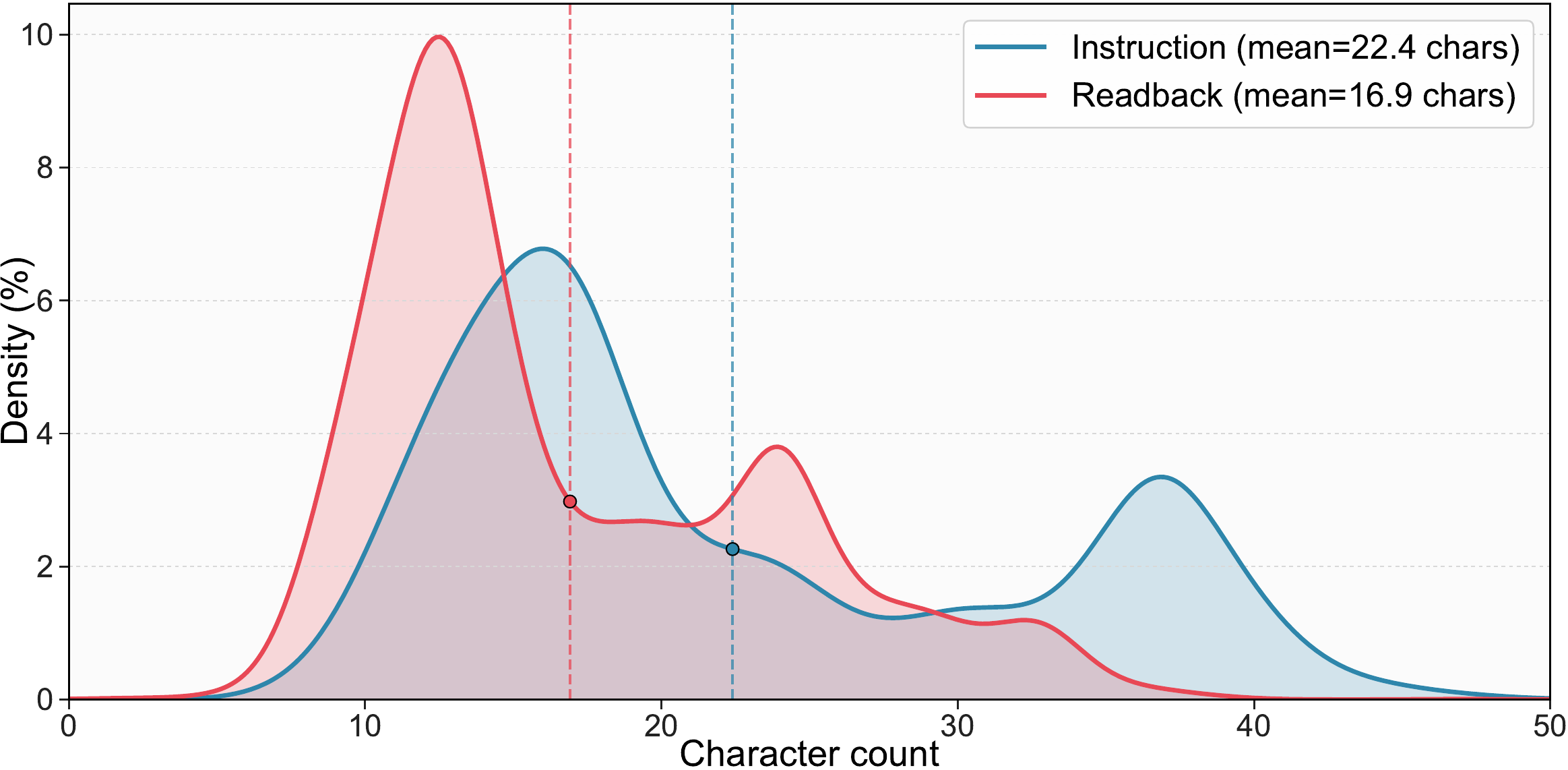}
    \caption{Length distributions of ATCo instructions and pilot readbacks.}
    \label{fig:instruction_readback_length_distribution}
\end{figure}

We perform preliminary analyses on the dataset. As shown in \hyperref[fig:silhouette]{Fig. 9}, the silhouette coefficients~\citep{rousseeuw1987silhouettes} across all five classes are generally close to zero, indicating weak separability in the representation space derived from original features. Among these classes, Unknown exhibits the poorest clustering quality, with more negative values and stronger overlap with known classes.
This suggests that Unknown lacks stable semantic coherence and is unsuitable for conventional closed-set classification. \hyperref[fig:instruction_readback_length_distribution]{Fig. 10} shows the length distributions of ATCo instructions and pilot readbacks. Instructions are generally longer and contain complete operational intent with key parameters, whereas readbacks are shorter confirmations that preserve only safety-critical information through simplified wording. For ATCo instructions, a secondary peak around 37 characters further reflects the presence of complex instructions such as compound commands and procedural clearances.
Because readbacks are often compressed, omissions, substitutions, and non-standard expressions may remain difficult to identify from surface form alone. This asymmetry makes readback monitoring inherently challenging.

\subsubsection{Parameters}

For dense feature extraction in open-set detection, the plug-in encoder $f(\cdot;\theta)$ is instantiated with the pretrained Erlangshen-DeBERTa-v2-97M-Chinese model\footnote{For more information: \url{https://huggingface.co/IDEA-CCNL/Erlangshen-DeBERTa-v2-97M-Chinese}}, with a hidden dimension $d=768$ and a dropout rate of 0.2. The POC is trained for 30 epochs with a batch size of 8, a maximum sequence length of 128, and a learning rate of $1\times 10^{-5}$, optimized by AdamW~\citep{loshchilov2017decoupled}. After training, its parameters are frozen and the model is used as a plug-in during LLM inference. The random seed is fixed to 42 to ensure reproducibility.
For DEAR, $\mathfrak{h}(\cdot)$ is represented using TF-IDF features~\citep{salton1988term}, as ATC communications are highly standardized and dominated by recurring domain-specific expressions. Character-level TF-IDF is applied separately to ATCo instructions and pilot readbacks, retaining up to 10,000 TF-IDF features. To assess both zero-shot and few-shot capability, the number of examples per class is varied from $N=0$ to $N=4$, and the candidate pool size is defined as $M=\rho N$. The intent and slot annotations attached to the examples are derived from our previous ATC semantic understanding model~\citep{ZHANG2026103812}.
Three key hyperparameters are determined via grid search, including the number of nearest neighbors $k$, the pool ratio $\rho$, and the MMR relevance and diversity trade-off coefficient $\alpha$. Specifically, $k$ is searched over $\{1,2,3,\dots,10\}$, $\rho$ over $\{3,4,5,\dots,10\}$, and $\alpha$ over $\{0,0.1,0.2,\dots,1.0\}$. The best-performing configuration, $k=8$, $\rho=9$, and $\alpha=0.3$, is used in all subsequent experiments. All LLM inference is conducted within the LLaMA-Factory framework~\citep{zheng2024llamafactory}, accelerated by the vLLM backend~\citep{kwon2023efficient} for high-throughput decoding. All experiments are performed on a single NVIDIA Tesla L40S GPU with 48\,GB of memory.

\subsubsection{Baselines}

\textbf{Readback Anomaly Detection Baselines.}
We compare against eight baselines spanning rule-based, supervised, and ICL paradigms.
\textsf{Regex}~\citep{prasad2021grammar} applies handcrafted grammar rules and regular expressions over transcribed ATCo--pilot communication pairs to detect character discrepancies between instructions and readbacks.
\textsf{N-gram}~\citep{helmke2021readback} adapts character-level F1, averaged over 1 to 3 grams, to measure information overlap between an instruction and its readback, classifying pairs based on empirically tuned coverage thresholds.
\textsf{DeBERTa}~\citep{he2021deberta} fine-tunes a DeBERTa encoder on the open-set readback classes, providing a supervised learning baseline and serving as the backbone encoder of our POC module.
\textsf{MilTOOD}~\citep{DarrinSGCPC24} performs open-set detection by aggregating cosine-similarity anomaly scores across all transformer layers rather than relying only on the final-layer representation.
\textsf{Simple ICL}~\citep{brown2020language} retrieves the most similar training examples within each label bucket through cosine similarity and presents them as in-context examples.
\textsf{DICL}~\citep{KapuriyaKGB25} uses dense embeddings and applies MMR within each label bucket to promote example diversity.
\textsf{SuperICL}~\citep{XuXWLZM24} augments top $k$ retrieval by injecting each example's small model prediction and confidence score into the prompt, thereby providing the LLM with a soft prior at inference time.
\textsf{GenICL}~\citep{zhang2025learning} trains a lightweight multi-layer perceptron reranker on pairwise preferences elicited from an LLM to select examples.

\textbf{Semantic Understanding Baselines.}
For the slot filling and intent recognition subtasks, we evaluate five baselines.
\textsf{BiLSTM-CRF}~\citep{LampleBSKD16} frames slot filling as sequence labeling with a bidirectional LSTM and a Conditional Random Field (CRF) decoding layer, while intent recognition is implemented through an additional sentence-level softmax classifier over the BiLSTM representations.
\textsf{JointBERT}~\citep{DBLP:journals/corr/abs-1902-10909} fine-tunes BERT to jointly predict slot and intent labels in a single forward pass through shared representations.
\textsf{HIN}~\citep{ZHANG2026103812} proposes a multitask hierarchical attention network specifically designed for joint slot and intent understanding, exploiting the structural hierarchy of ATCo--pilot communications.
\textsf{Aligner\textsuperscript{2}}~\citep{zhu2024aligner2} enhances joint multiple slot filling and intent recognition through adjustive and forced cross-task alignment modules that explicitly propagate intent signals into slot prediction.
\textsf{ELSF}~\citep{ZhuHHXLLCXY24} introduces an entity-level slot filling framework that decomposes slot filling into entity boundary detection and type assignment, jointly optimized with a multi-intent detection objective.

\subsubsection{Evaluation Metrics}
For a test set of $N$ samples, Accuracy and F1 are defined as
\begin{equation}
\mathrm{Acc}=\frac{1}{N}\sum_{i=1}^{N}\mathbb{I}(\hat{y}_i=y_i),\qquad
\mathrm{F1}=\frac{2\cdot\mathrm{P}\cdot\mathrm{R}}{\mathrm{P}+\mathrm{R}},
\end{equation}
where $y_i$ and $\hat{y}_i$ denote the ground truth and predicted labels of the $i$-th sample, and $\mathbb{I}(\cdot)$ is the indicator function that equals 1 if the enclosed condition holds and 0 otherwise. Precision $(\mathrm{P})$ measures the proportion of correctly predicted positive samples among all predicted positives, i.e., $\mathrm{P}=\mathrm{TP}/(\mathrm{TP}+\mathrm{FP})$. Recall $(\mathrm{R})$ measures the proportion of correctly predicted positives among all actual positives, i.e., $\mathrm{R}=\mathrm{TP}/(\mathrm{TP}+\mathrm{FN})$, where TP, FP, and FN denote the numbers of true positives, false positives, and false negatives, respectively. F1 is the harmonic mean of Precision and Recall, providing a balanced measure under class imbalance.

For open-set readback anomaly detection, we further report kF1 and uF1, which apply the above F1 formulation to the known and unknown classes, respectively, together with the harmonic mean (HM)
\begin{equation}
\mathrm{HM}=\frac{2\cdot \mathrm{kF1}\cdot \mathrm{uF1}}{\mathrm{kF1}+\mathrm{uF1}}.
\end{equation}

For semantic understanding, we reuse Accuracy and F1 as Intent Acc and Slot F1, computed over predicted intents $\hat{\mathrm{i}}_i$ and slots $\hat{\mathrm{s}}_i$, respectively. In addition, Semantic Frame Accuracy (SFA) is adopted as a strict joint metric,
\begin{equation}
\mathrm{SFA}=\frac{1}{N}\sum_{i=1}^{N}\mathbb{I}(\hat{\mathrm{i}}_i=\mathrm{i}_i)\cdot\mathbb{I}(\hat{\mathrm{s}}_i=\mathrm{s}_i),
\end{equation}
where a sample is counted as correct only if both its intent and all slots are predicted correctly.

\subsection{Experimental Result}

\subsubsection{Baseline Comparison}

\begin{table}[htbp]
\centering
\caption{Experimental results (\%) on the APCP dataset using different comparison methods.}
\label{tab:main_results}
\footnotesize
\setlength{\tabcolsep}{4pt}
\setlength{\cmidrulewidth}{0.4pt}
\begin{tabular*}{\textwidth}{@{\extracolsep{\fill}} ll ccccccccc S[table-format=2.2, detect-weight=true]}
\toprule
\multicolumn{2}{c}{\textit{Type}}
  & \multicolumn{2}{c}{\textit{Rule-based}}
  & \multicolumn{2}{c}{\textit{Supervised Learning}}
  & \multicolumn{6}{c}{\textit{LLM-based ICL}(Qwen3-14B)} \\
\cmidrule(lr){1-2}\cmidrule(lr){3-4}\cmidrule(lr){5-6}\cmidrule(lr){7-12}
\multicolumn{2}{c}{Method} & Regex & N-gram & DeBERTa & MilTOOD & Manual & ICL & DICL & GenICL & SuperICL & \textbf{SCOPE} \\
\midrule
\multirow{2}{*}{Overall}
  & Acc & 25.61 & 25.96 & 87.02 & 78.33 & 42.63 & — & — & — & — & — \\
  & F1   & 25.44 & 25.46 & 86.79 & 79.08 & 50.11 & — & — & — & — & — \\
\hdashline\addlinespace[3pt]
\multirow{2}{*}{0-shot}
  & Acc & — & — & — & — & — & 47.98 & 47.98 & 47.98 & \underline{66.23} & \textbf{86.69} \\
  & F1   & — & — & — & — & — & 31.12 & 31.12 & 31.12 & \underline{62.61} & \textbf{86.25} \\[4pt]
\multirow{2}{*}{1-shot}
  & Acc & — & — & — & — & — & 65.61 & 67.98 & 65.88 & \underline{80.88} & \textbf{88.51} \\
  & F1   & — & — & — & — & — & 61.52 & 63.79 & 61.46 & \underline{79.75} & \textbf{88.53} \\[4pt]
\multirow{2}{*}{2-shot}
  & Acc & — & — & — & — & — & 66.05 & 69.47 & 66.93 & \underline{80.88} & \textbf{89.56} \\
  & F1   & — & — & — & — & — & 63.05 & 66.25 & 63.64 & \underline{79.77} & \textbf{89.54} \\[4pt]
\multirow{2}{*}{3-shot}
  & Acc & — & — & — & — & — & 67.81 & 70.96 & 66.58 & \underline{83.86} & \textbf{90.61} \\
  & F1   & — & — & — & — & — & 65.41 & 67.96 & 62.92 & \underline{83.19} & \textbf{90.58} \\[4pt]
\multirow{2}{*}{4-shot}
  & Acc & — & — & — & — & — & 65.70 & 67.19 & 64.74 & \underline{84.12} & \textbf{91.05} \\
  & F1   & — & — & — & — & — & 66.30 & 63.36 & 60.24 & \underline{83.61} & \textbf{91.01} \\
\bottomrule
\end{tabular*}
\vspace{2pt}
\raggedright
\footnotesize{\textit{Note:} ``—'' indicates not applicable. \textbf{Bold} denotes the best performance, and \underline{underlining} denotes the second-best.}
\end{table}

\hyperref[tab:main_results]{Table 3} compares different methods across three paradigms: rule-based matching, supervised learning, and LLM-based ICL. Overall, with Qwen3-14B~\citep{yang2025qwen3} as the backbone, SCOPE achieves the best performance across all evaluation settings. This result highlights the advantage of combining open-set priors from the plug-in classifier, example retrieval that balances relevance and diversity, and structured semantic reasoning.
The rule-based methods perform poorly, showing that readback monitoring cannot rely only on fixed lexical rules. Furthermore, supervised models clearly improve over rule-based methods, but their performance remains constrained by the difficulty of the open-set setting.

Among LLM-based approaches, manual prompting yields only 42.63\% Acc and 50.11\% F1, showing that detailed task descriptions alone are insufficient to cover all conditions in real ATCo--pilot communications. At 0-shot, ICL, DICL, and GenICL achieve the same results because no examples are provided.
As the number of shots increases, all ICL methods show consistent improvement, reflecting that more examples provide increasingly useful demonstrations for the LLM. With four examples per class, SuperICL achieves the strongest baseline performance due to its ability to construct more informative in-context examples. Nevertheless, its performance remains substantially lower than that of SCOPE, with a 6.93\% gap in accuracy. These results suggest that SCOPE benefits from more effective example guidance and semantic alignment, which together produce more reliable reasoning for ATC readback monitoring.

\subsubsection{Plug-in and LLM Synergy}

\begin{figure}[t!]
    \centering
    \includegraphics[width=0.75\linewidth]{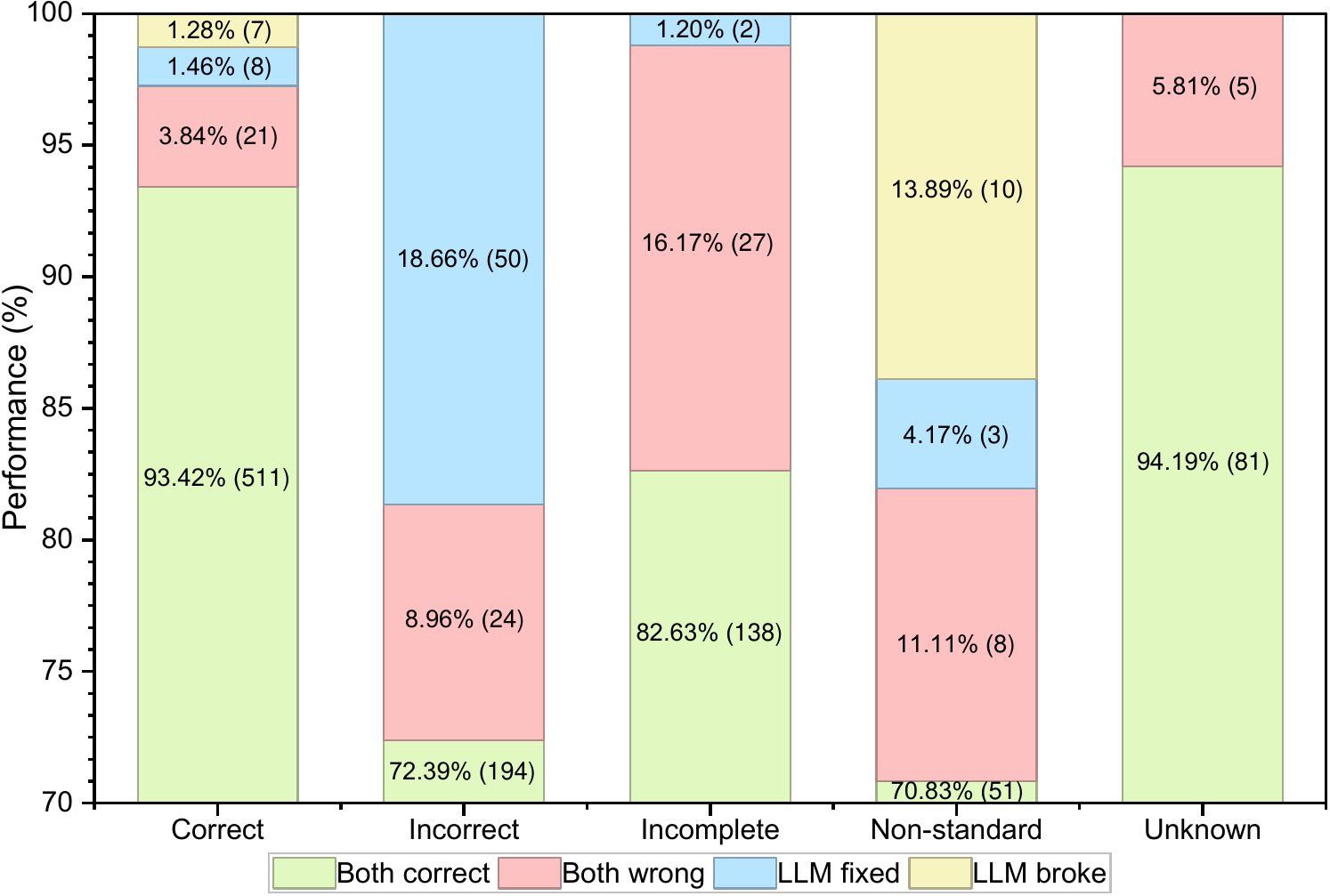}
    \caption{Class distribution of collaborative outcomes between the plug-in model and the LLM label refinement.}
    \label{fig:synergy}
\end{figure}

We further analyze how the plug-in model (POC) and the LLM interact during the final prediction stage. The frozen plug-in model first provides an initial label prior based on open-set classification, while the LLM subsequently refines this prior through DEAR and ATCoT. To examine whether this refinement is beneficial or harmful, we divide the samples into four collaborative outcomes: samples correctly predicted by both models, samples misclassified by both models, samples corrected by the LLM, and samples degraded by the LLM.
As shown in \hyperref[fig:synergy]{Fig. 11}, each stacked bar corresponds to one ground truth class, namely \emph{correct}, \emph{incorrect}, \emph{incomplete}, \emph{non-standard}, and \emph{unknown}. The segments within each bar indicate the proportions and counts of the four collaborative outcomes.
The results show that the LLM corrects substantially more plug-in errors than it introduces. After LLM refinement, the total accuracy and F1 increase from 87.02\% and 86.79\% to 91.05\% and 91.01\%, corresponding to gains of 4.03\% and 4.22\%, respectively. The improvement is mainly driven by the \emph{incorrect} class, whose accuracy rises from 72.39\% to 91.05\%. Many samples initially predicted as \emph{correct} by the plug-in model are reassigned to \emph{incorrect} after LLM reasoning, indicating that the LLM is effective at identifying subtle mismatches within semantically similar readbacks. In contrast, the \emph{unknown} class remains largely unchanged, suggesting that the plug-in model already provides a stable open-set boundary.
These results indicate a complementary relationship between the two components. The plug-in model provides label boundaries, while the LLM provides fine-grained semantic refinement by combining representative demonstrations with explicit intent and slot consistency checking. The synergy enables SCOPE to correct subtle readback errors without sacrificing its ability to recognize unknown communication patterns.

\subsubsection{Ablation Study}

\begin{table}[htbp]
\centering
\caption{Ablation experiments of different modules. $\downarrow$ indicates the performance drop relative to the full SCOPE.}
\label{tab:ablation}
\begin{tabular}{c c c c c}
\toprule
\multirow{2}{*}{No.} & \multirow{2}{*}{Module} & \multirow{2}{*}{Variants} & \multicolumn{2}{c}{4-shot (\%)} \\
\cmidrule(lr){4-5}
& & & Acc & F1 \\
\midrule
1 & POC & w/o POC & 73.77 ($\downarrow$17.28) & 74.22 ($\downarrow$16.79) \\
\midrule
2 & \multirow{3}{*}{DEAR}
  & w/o Anchored       & 89.30 ($\downarrow$1.75) & 89.19 ($\downarrow$1.82) \\
3 & 
  & w/o Diversity        & 88.51 ($\downarrow$2.54) & 88.45 ($\downarrow$2.56) \\
4 & 
  & Random order       & 88.16 ($\downarrow$2.89) & 88.11 ($\downarrow$2.90) \\
\midrule
5 & \multirow{2}{*}{ATCoT}
  & w/o examples semantics           & 82.72 ($\downarrow$8.33) & 82.15 ($\downarrow$8.86) \\
6 & 
  & w/ test semantics & 84.65 ($\downarrow$6.40) & 84.55 ($\downarrow$6.46) \\
\hdashline\addlinespace[3pt]
  & SCOPE & Full model & \textbf{91.05} & \textbf{91.01} \\
\bottomrule
\end{tabular}
\end{table}

To investigate the contribution of each module, we conduct ablation experiments under six variant settings. For clarity, the variants are defined as follows:
\begin{itemize}
    \item \textbf{Variant 1} removes the POC, so the plug-in predicted label is excluded from both the examples and the test prompt.
    \item \textbf{Variant 2} removes the candidate-pool construction stage and retrieves examples with the MMR re-ranking stage only.
    \item \textbf{Variant 3} removes the MMR re-ranking stage and retrieves examples with the candidate-pool construction stage only.
    \item \textbf{Variant 4} keeps the retrieved examples but randomizes the order in which they are presented to the LLM.
    \item \textbf{Variant 5} removes annotated semantic information from examples.
    \item \textbf{Variant 6} retains semantic annotations in the examples and directly provides the annotated semantics of the test sample, instead of requiring the LLM to infer them.
\end{itemize}

\hyperref[tab:ablation]{Table 4} shows that removing any module leads to performance degradation, demonstrating the necessity of each module. Among all modules, the POC contributes the most, as removing it in Variant 1 causes the largest degradation, with both Acc and F1 dropping by more than 16\%. This result indicates that the plugin provides a strong initial label prior, allowing the LLM to refine predictions from a plausible starting point rather than reasoning from raw utterances alone.

For DEAR, Variants 2 to 4 all perform worse than the full model, with Variant 4 yielding the weakest retrieval performance. This performance shows that effective example construction depends not only on semantic relevance, but also on scenario consistency, diversity, and example order.

ATCoT also provides a substantial gain. Removing semantics in Variant 5 results in a large drop, indicating that structured semantic cues are essential for LLM reasoning.
It is noted that variant 6 provides explicit semantics for the test sample, but performance degrades because these inputs can interfere with the model’s own reasoning. The full model performs better because semantics are used as intermediate guidance rather than fixed inputs.

\subsubsection{Explanation and Correction}
\label{sec:exp_explanation_correction}

To move the proposed framework closer to a deployable ATC decision-support prototype, this subsection evaluates two downstream outputs built on readback anomaly detection: the \emph{explanation} for each detected ATCo--pilot communication pair and the \emph{correction} that generates a corrected readback when the pilot's utterance is judged anomalous. Explanation quality is assessed by strong external LLM judges, while correction quality is evaluated against human-annotated ground truth.

1) \textit{Explanation.}
Because the explanations are generated by an LLM without reference explanations, reference-based metrics are not suitable for evaluation. Following the LLM-as-a-Judge method, we employ two strong frontier commercial models, GPT-5.3~\citep{openai2026gpt53instantcard} and Claude Opus 4.7~\citep{anthropic2026opus47}, as independent judges for reasoning-intensive assessment.
Each judge is given the instruction, readback, predicted label, and explanation, and is asked to rate the explanation from 1 to 5 along three dimensions. Completeness measures whether all inconsistent slots and their operational implications are covered. Accuracy evaluates whether the identified discrepancies match the actual error pattern. Conciseness reflects whether the explanation remains clear and free of redundancy under operational time constraints.

\begin{table}[htbp]
\centering
\caption{LLM-as-a-Judge scores for explanations, averaged over the test set (rate from 1 to 5, higher is better).}
\label{tab:model_metrics}
\begin{tabular}{lccc}
\toprule
Judge Model & Completeness & Accuracy & Conciseness \\
\midrule
GPT-5.3           & 4.24 & 4.66 & 4.13 \\
Claude Opus 4.7   & 3.87 & 4.81 & 4.20 \\
\bottomrule
\end{tabular}
\end{table}

The averaged results over the test set are reported in \hyperref[tab:model_metrics]{Table 5}. Both judges assign high Accuracy scores of 4.66 and 4.81 out of 5, showing that the generated explanations are generally consistent with the actual error patterns rather than hallucinated content. Completeness and Conciseness are slightly lower but still remain above 4.0 in most settings, except for Claude's Completeness score of 3.87. This result suggests a moderate trade-off between detailed explanations and concise outputs, which aligns with the ATC preference for short and focused alerts. The similar Accuracy scores from two independent judges also indicate that the evaluation results are relatively stable across different LLM evaluators.

\begin{figure}[h!]
    \centering
    \includegraphics[width=\linewidth]{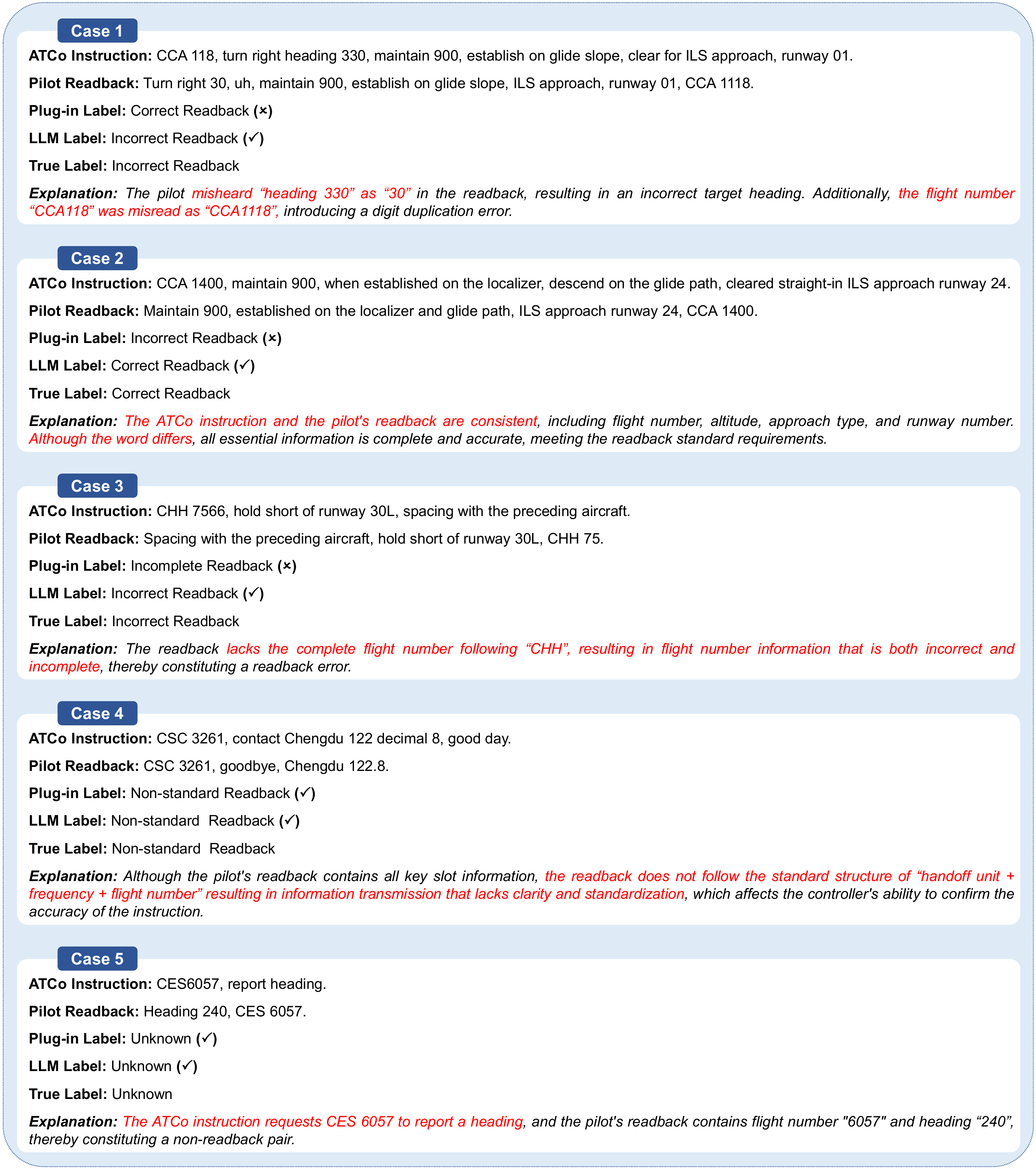}
    \caption{Representative cases across all classes. Each case shows ATCo instruction, pilot readback, plug-in label, LLM label, ground truth, and generated explanation.}
    \label{fig:explanation_case}
\end{figure}

\hyperref[fig:explanation_case]{Fig. 12} illustrates five representative cases covering all classes in ATCo--pilot communications. The first three cases are correctly classified by the LLM but misclassified by the plug-in, each reflecting a different failure mode.
In Case~1, the plug-in falsely classifies an \emph{Incorrect} readback as \emph{Correct}. This is likely because the DeBERTa encoder maps utterances into a dense embedding space, where samples with small numerical differences may remain close to correct readbacks. SCOPE corrects the plug-in prediction and explains the error by explicitly identifying the faulty slot, making the decision interpretable to operators.
Case~2 shows the opposite situation, where the plug-in incorrectly predicts an \emph{Incorrect} label for a valid \emph{Correct} readback. The pilot intentionally omits non-essential words, but the plug-in lacks sufficient semantic understanding to recognize this operationally acceptable abbreviation. The LLM correctly identifies the readback label with ATC knowledge and high-quality examples.
Case~3 is likely caused by upstream ASR segmentation, where the flight number ``7566'' is truncated to ``75''. The plug-in interprets this corrupted value as an \emph{Incomplete}, whereas SCOPE correctly recognizes it as an \emph{Incorrect} substitution error. This result reflects the operational principle that even small numerical deviations in safety-critical slots must be treated as errors.
Cases~4 and~5 show both classifiers producing consistent predictions on structurally clear inputs, including a \emph{Non-standard} readback and an open-set \emph{Unknown} communication. The consistent predictions and faithful explanations suggest that SCOPE can effectively handle both readback anomalies and unseen ATCo--pilot communications without weakening the plug-in's existing strengths.

2) \textit{Correction.}
As shown in \hyperref[tab:correction_metrics]{Table 6}, evaluation is conducted on 505 anomalous samples, including all communications outside the \emph{Correct} Readback class within the test set. Due to phraseology regularization, the correction module follows standard ATC communication rules and can reliably reconstruct corrected readbacks once an anomaly is successfully detected by the upstream anomaly detection module. Because a small number of anomalous samples are misclassified, the overall correction rate reaches 96.63\%.
Among different classes, \emph{Incomplete} Readback achieves perfect correction performance at 100.00\%, indicating that all incomplete cases are successfully detected and reconstructed. \emph{Incorrect} Readback achieves the lowest correction rate of 94.72\%, suggesting that semantically similar but safety-critical substitutions remain the most challenging cases for detection. These results demonstrate the effectiveness of combining LLM semantic reasoning with rule-based reconstruction. The LLM identifies semantic anomalies and recovers missing operational information, while the rule-guided generation enforces strict compliance with ATC phraseology.

\begin{table}[htbp]
\centering
\caption{Analysis of the correction performance.}
\label{tab:correction_metrics}
\begin{tabular}{lcc}
\toprule
Metric & Value (\%) & Corrected / Total \\
\midrule
Overall correction rate             & 96.63 & $488/505$ \\
\quad $\bullet$ Non-standard Readback           & 96.67 & $58/60$ \\
\quad $\bullet$ Incomplete Readback             & 100.00 & $161/161$ \\
\quad $\bullet$ Incorrect Readback               & 94.72 & $269/284$ \\
\bottomrule
\end{tabular}
\end{table}

\subsubsection{Open-set Analysis}

\begin{figure}[h!]
    \centering
    \begin{subfigure}[t]{0.48\linewidth}
        \centering
        \includegraphics[width=\linewidth]{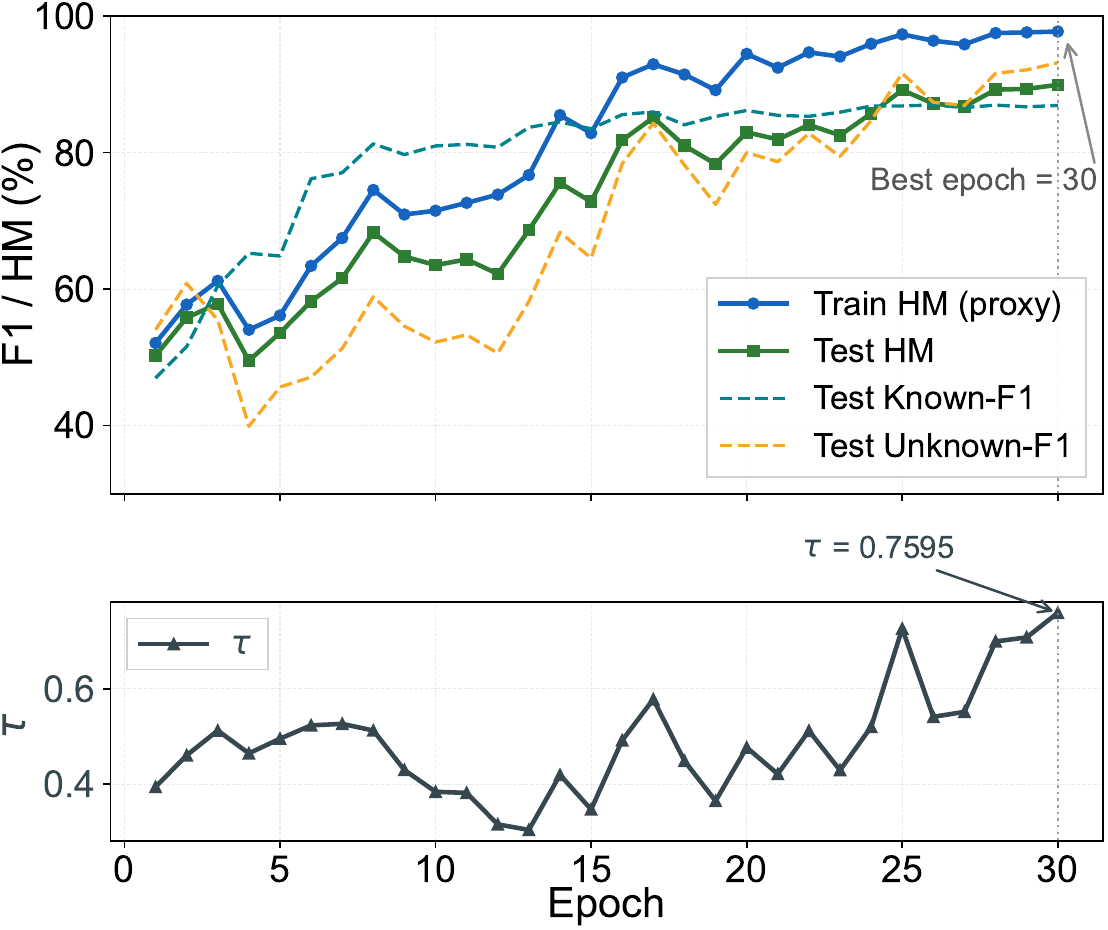}
        \caption{Training dynamics of HM, Known/Unknown-F1, and \(\tau\).}
        \label{fig:epoch_metrics}
    \end{subfigure}
    \hfill
    \begin{subfigure}[t]{0.50\linewidth}
        \centering
        \includegraphics[width=\linewidth]{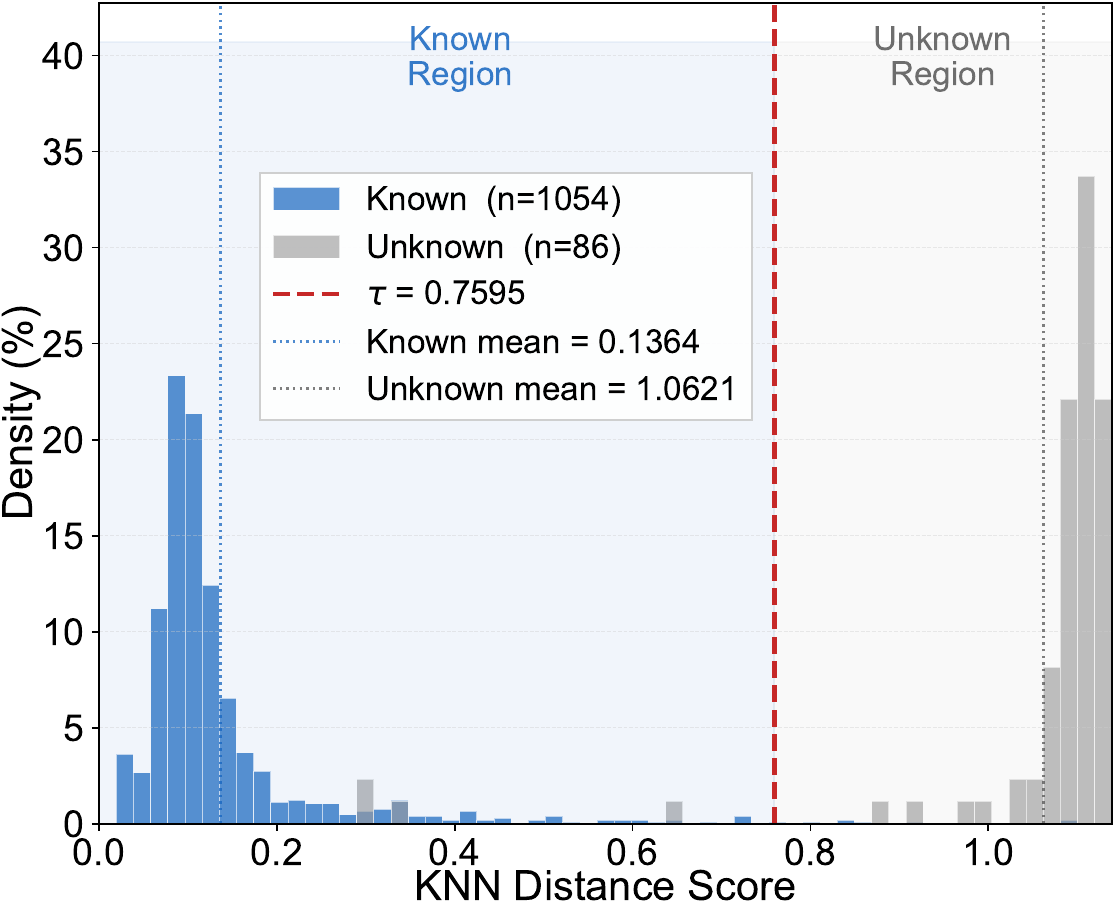}
        \caption{KNN distance distribution.}
        \label{fig:knn_distribution}
    \end{subfigure}
    \caption{Quantitative evaluation of POC for open-set readback recognition.}
    \label{fig:poc_quantitative}
\end{figure}

To analyze the role of the POC in open-set readback anomaly detection, we conduct both quantitative and qualitative evaluations. First, we record the training and test HM scores, together with the calibrated KNN threshold \(\tau\), at each epoch. As shown in \hyperref[fig:epoch_metrics]{Fig. 13(a)}, the training and testing HM scores generally follow a consistent upward trend, indicating that the learned representation improves open-set discrimination without overfitting. Since \(\tau\) is selected according to the KNN distance distribution rather than directly optimized by gradient descent, its fluctuation reflects the evolving feature-space boundary between known and unknown samples. Notably, the test HM decreases from epoch 8 to epoch 12, mainly because the Unknown-F1 drops during this period. This suggests that the model first strengthens known-class discrimination, but some unknown samples close to known-class boundaries are temporarily absorbed into known regions before the open-set boundary becomes stable.
\hyperref[fig:knn_distribution]{Fig. 13(b)} further shows the KNN distance distribution at the best epoch. Known samples are concentrated in the low-distance region, with a mean distance of 0.1364, whereas unknown samples are mainly distributed in the high-distance region, with a mean distance of 1.0621. The calibrated threshold \(\tau=0.7595\) lies between the two distributions and separates most known and unknown samples with only limited overlap. This indicates that POC learns a feature space with good open-set separability, where unknown communications can be detected by their distance from known readback regions.

\begin{figure}[h!]
    \centering
    \begin{subfigure}[t]{0.48\linewidth}
        \centering
        \includegraphics[width=\linewidth]{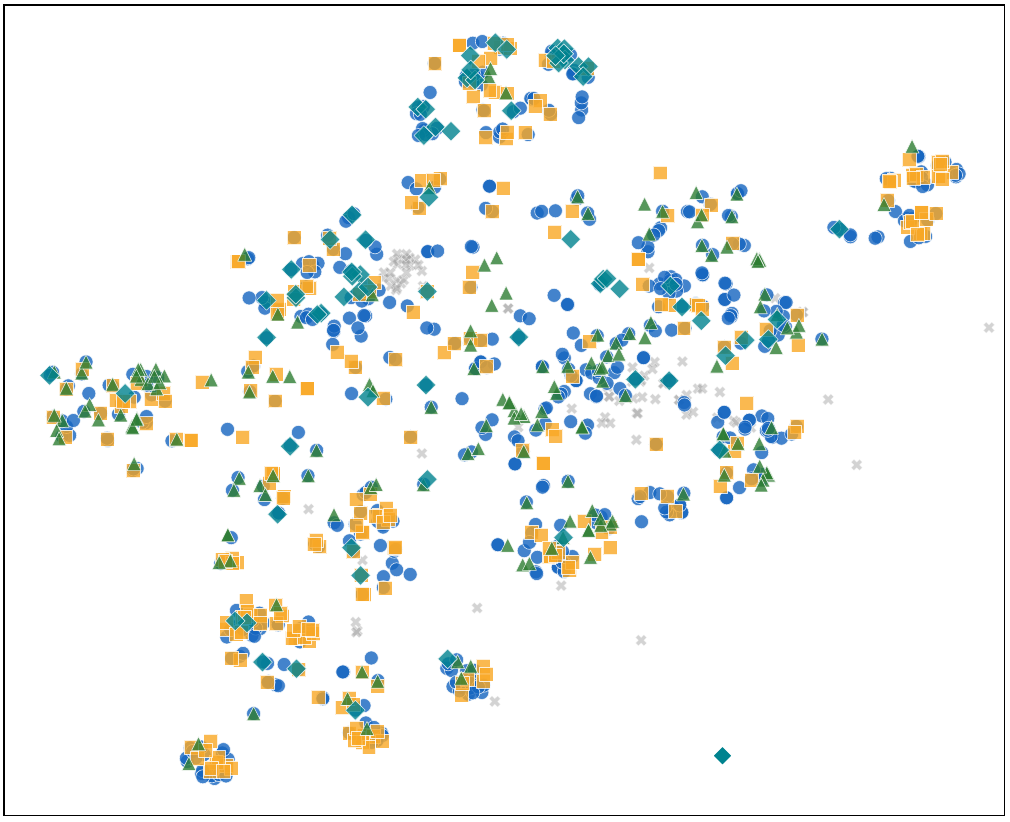}
        \caption{Before POC training.}
        \label{fig:tsne_before}
    \end{subfigure}
    \hfill
    \begin{subfigure}[t]{0.48\linewidth}
        \centering
        \includegraphics[width=\linewidth]{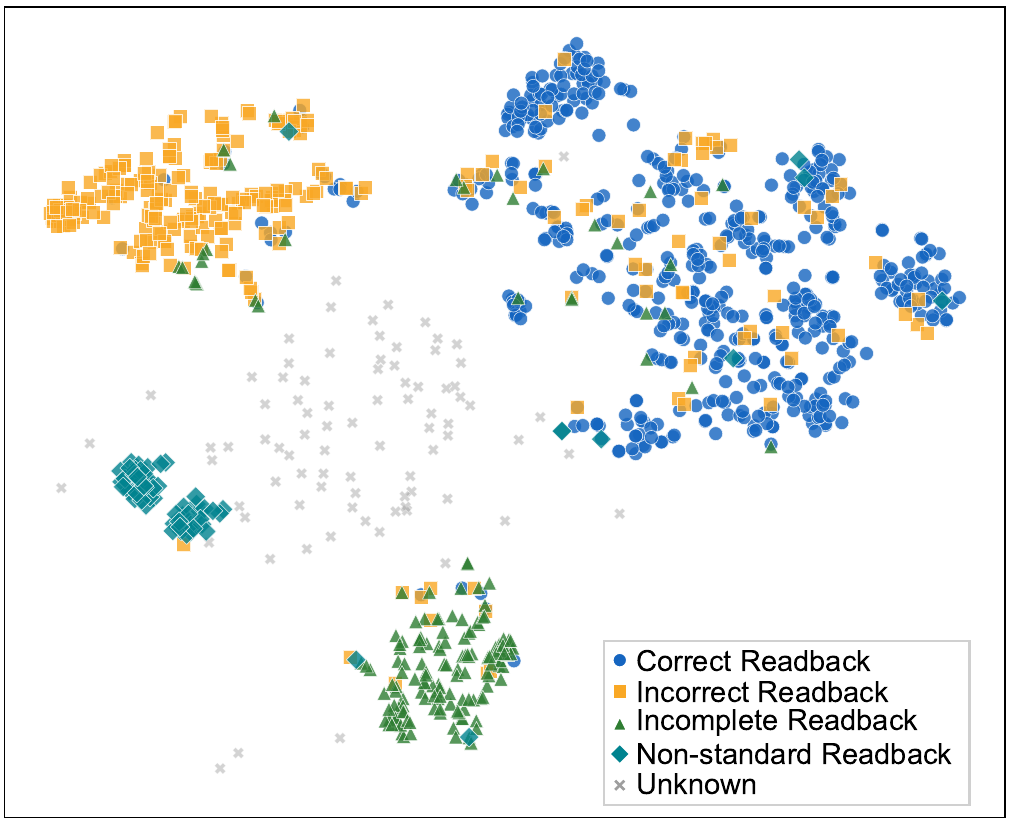}
        \caption{After POC training.}
        \label{fig:tsne_after}
    \end{subfigure}
    \caption{t-SNE visualization of the feature space before and after POC training.}
    \label{fig:poc_tsne}
\end{figure}

To provide a more intuitive view, we visualize the feature space before and after POC training using t-SNE in a two-dimensional space. As shown in \hyperref[fig:tsne_before]{Fig. 14(a)}, before training, all readback classes are highly mixed. After training, \hyperref[fig:tsne_after]{Fig. 14(b)} shows that known readback classes form more compact and distinguishable regions, while unknown samples are mostly mapped outside the known class clusters with a scattered distribution. This pattern is consistent with the open-set setting, where unknown communications do not share a stable class prototype. It further indicates that POC learns semantic boundaries among known readback states and enables unseen communication patterns in the test set to be detected through feature space deviation.

Nevertheless, both the quantitative results and the visualization in \hyperref[fig:tsne_after]{Fig. 14(b)} indicate that some known classes, especially \emph{Correct} and \emph{Incorrect} Readback, are still not fully separable in the learned feature space. This is because readback errors are often triggered by subtle changes in safety-critical slots, such as a single altitude, heading, runway, or callsign, while the overall sentence semantics may remain highly similar. Therefore, POC is not used as a standalone final classifier. Instead, it serves as an explicit open-set prior that provides initial predictions, allowing the LLM to further perform reasoning and make predictions that are more consistent with operational semantics.

\subsubsection{In-Context Example Retrieval}
To examine how different retrieval strategies affect SCOPE, we replace the DEAR module with alternative retrieval methods while keeping the rest of the framework unchanged. Unlike the baseline comparison in \hyperref[tab:main_results]{Table 3}, where each baseline is evaluated within its original framework, here all retrieval methods are placed inside SCOPE, isolating the effect of retrieval choice from other architectural differences.

\begin{table}[h!]
\centering
\caption{Performance (\%) of different retrieval strategies for in-context example selection on the APCP dataset.}
\label{tab:retrieval_strategy}
\small
\setlength{\tabcolsep}{6pt}
\setlength{\cmidrulewidth}{0.4pt}
\sisetup{detect-weight=true, detect-shape=true, detect-mode=true, mode=text}
\begin{tabular}{ll *{5}{S[table-format=2.2]}}
\toprule
\multicolumn{2}{c}{LLM type}
  & \multicolumn{5}{c}{Qwen3-14B} \\
\cmidrule(lr){1-2}\cmidrule(lr){3-7}
\multicolumn{2}{c}{Method} & {ICL} & {DICL} & {GenICL} & {SuperICL} & {\textbf{SCOPE}} \\
\midrule
\multirow{2}{*}{0-shot}
  & Acc & 86.69 & 86.69 & 86.69 & 86.69 & 86.69 \\
  & F1   & 86.25 & 86.25 & 86.25 & 86.25 & 86.25 \\[4pt]
\multirow{2}{*}{1-shot}
  & Acc & \underline{88.33} & 82.54 & 82.72 & 85.26 & \bfseries 88.51 \\
  & F1   & \underline{88.30} & 82.21 & 82.62 & 85.35 & \bfseries 88.53 \\[4pt]
\multirow{2}{*}{2-shot}
  & Acc & \underline{87.89} & 83.68 & 82.98 & 86.40 & \bfseries 89.56 \\
  & F1   & \underline{87.91} & 83.71 & 83.05 & 86.66 & \bfseries 89.54 \\[4pt]
\multirow{2}{*}{3-shot}
  & Acc & \underline{88.42} & 84.39 & 82.37 & 85.70 & \bfseries 90.61 \\
  & F1   & \underline{88.45} & 84.40 & 82.49 & 85.95 & \bfseries 90.58 \\[4pt]
\multirow{2}{*}{4-shot}
  & Acc & \underline{88.42} & 85.00 & 83.68 & 86.67 & \bfseries 91.05 \\
  & F1   & \underline{88.31} & 84.83 & 83.70 & 86.88 & \bfseries 91.01 \\
\bottomrule
\end{tabular}
\end{table}

\hyperref[tab:retrieval_strategy]{Table 7} shows that at 0-shot, all methods share an identical prompt, yielding the same performance of 86.69\% Acc and 86.25\% F1, since no examples are provided. When examples are introduced, performance differences become evident. By 4-shot, SCOPE achieves 91.05\% Acc and 91.01\% F1, outperforming the strongest baseline by 2.63\% and the weakest by 7.37\%. The performance gap increases steadily from 0.18\% at 1-shot to 2.63\% at 4-shot, indicating that performance is primarily determined by the quality of retrieved examples rather than the amount of context.

The relative performance of the baselines reveals the characteristics of readback retrieval. Vanilla ICL consistently ranks second and outperforms neural retrieval methods, suggesting that simple lexical similarity can better capture word discrepancies that are critical for readback monitoring. In contrast, retrieval methods based mainly on dense semantic similarity show weaker performance. DICL and SuperICL achieve 85.00\% and 86.67\% Acc at 4-shot, but remain below SCOPE because they neglect the asymmetric structure of ATCo instructions and pilot readbacks. These results demonstrate that effective example retrieval for ATC monitoring requires both scenario consistency and class-discriminative information, which validates the design of the proposed DEAR module.

\subsubsection{Semantic Understanding Auxiliary Task}

\begin{table}[h!]
\centering
\caption{Comparison of models on slot filling and intent recognition.}
\label{tab:results}
\begin{tabular}{lcccccc}
\toprule
Metric (\%)& BiLSTM-CRF & JointBERT & HIN & Aligner\textsuperscript{2} & ELSF & \textbf{Ours} \\
\midrule
Intent Acc & 90.53 & 87.72 & 84.65 & 91.71 & \underline{93.82} & \textbf{95.22} \\
Slot F1    & 90.70 & \underline{92.17} & \textbf{93.19} & 91.18 & 89.62 & 91.42 \\
SFA        & 63.51 & 64.21 & 64.87 & \underline{65.13} & 60.92 & \textbf{68.90} \\
\bottomrule
\end{tabular}
\end{table}

In the ATCoT module, the model not only predicts the final readback label but also infers intent and slot information. Intent reflects the operational purpose of an utterance, such as climb or descent instructions, while slots encode key parameters such as altitude, speed, or callsign. These elements are essential for structured understanding in ATCo--pilot communication.
To evaluate this capability, we conduct auxiliary tasks on slot filling and intent recognition, as shown in \hyperref[tab:results]{Table 8}. It should be noted that these tasks are introduced only as intermediate semantic cues rather than primary training objectives, and no task-specific optimization is applied. Even under this setting, the proposed method achieves the best Intent Acc and SFA, outperforming models specifically designed for joint slot filling and intent recognition.

This advantage can be attributed to the difference in modeling. Most baselines rely on BIO-style sequence labeling with token-level decisions, which is well-suited for precise boundary detection and thus leads to higher Slot F1. By contrast, the LLM performs semantic interpretation at the utterance level, jointly reasoning over intent and slot structures in a compositional manner. This global understanding reduces error propagation from local tagging decisions and improves overall semantic consistency.

\subsubsection{Backbone LLM Analysis}

To examine the impact of the backbone LLM on the readback anomaly detection task, we compare candidate models along three orthogonal dimensions: different training objectives within the same LLM family, different model families at the same parameter scale, and different parameter scales within the same family. All models are evaluated under 0 to 4-shot settings using accuracy and F1 as metrics.

\begin{figure}[t!]
    \centering
    \begin{subfigure}[t]{0.325\linewidth}
        \centering
        \includegraphics[width=\linewidth]{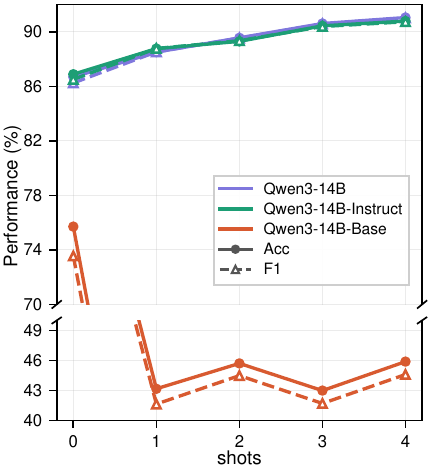}
        \caption{Same family, different purpose.}
        \label{fig:a}
    \end{subfigure}
    \hfill
    \begin{subfigure}[t]{0.325\linewidth}
        \centering
        \includegraphics[width=\linewidth]{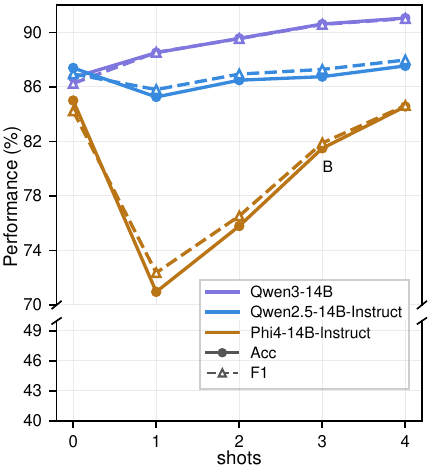}
        \caption{Different family, same size.}
        \label{fig:b}
    \end{subfigure}
    \hfill
    \begin{subfigure}[t]{0.325\linewidth}
        \centering
        \includegraphics[width=\linewidth]{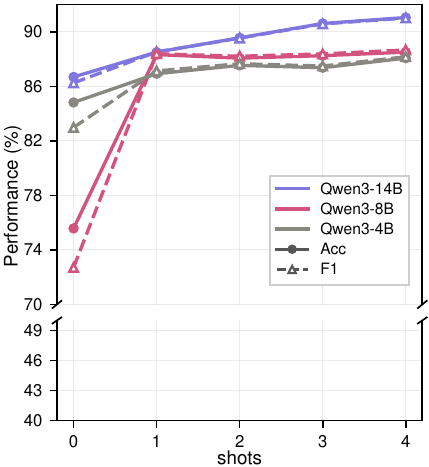}
        \caption{Same family, different size.}
        \label{fig:c}
    \end{subfigure}
    \caption{Performance comparison of different backbone LLMs under 0 to 4 shot settings. Solid lines denote accuracy and dashed lines denote F1.}
    \label{fig:abc}
\end{figure}

As shown in \hyperref[fig:a]{Fig. 15(a)}, Qwen3-14B and Qwen3-14B-Instruct exhibit nearly identical performance across all shot settings, with Qwen3-14B slightly outperforming at 4-shot with 91.05\% versus 90.79\%. This suggests that Qwen3-14B already possesses strong instruction-following ability, and additional instruction tuning provides only marginal benefit. In contrast, Qwen3-14B-Base drops sharply from 75.71\% at 0-shot to around 43–46\% once examples are introduced. Without instruction alignment, Qwen3-14B-Base fails to utilize in-context examples and is instead distracted by superficial patterns, indicating that the ICL task requires an instruction-tuned backbone.

\hyperref[fig:b]{Fig. 15(b)} compares three 14B-scale models. Qwen3-14B consistently surpasses Qwen2.5-14B-Instruct by about 3\% across all shot settings, reflecting stronger reasoning and instruction-following capability in the Qwen3 series. Phi4-14B-Instruct performs worst overall, particularly at 1-shot with 70.96\%, due to its weaker adaptation to Chinese inputs. Interestingly, it achieves a relatively high score of 85.00\% at 0-shot. In the absence of examples, the model relies heavily on the plug-in label in the prompt. Once examples are introduced, the mismatch between its pretraining data and the task domain introduces noise that degrades performance.

\hyperref[fig:c]{Fig. 15(c)} analyzes the effect of parameter scale within the Qwen3 family. From 1-shot onward, performance follows a clear scaling trend, with Qwen3-14B outperforming Qwen3-8B, which in turn exceeds Qwen3-4B. At 0-shot, however, Qwen3-4B reaches 84.82\% and exceeds Qwen3-8B at 75.58\%. This reflects a trade-off between model capacity and reliance on prompt signals. Smaller models tend to follow the plug-in label directly, which is advantageous when no examples are available. The 8B model has sufficient capacity to perform autonomous reasoning, which can lead to overinterpretation without exemplar guidance. Introducing a single example restores its performance to the expected trend.

\subsubsection{System Prototype}

To evaluate the applicability of SCOPE in safety-critical ATC scenarios, we further stratify ATCo--pilot communication pairs by operational risk. Ground operations, airborne control, and operational coordination are defined as high-risk communication categories because errors in these scenarios may directly affect flight safety and lead to severe operational consequences. In addition to accuracy and F1, we further report Recall, which measures the ability of the framework to correctly identify anomalous readbacks among real ATCo--pilot communications.

\begin{table}[h!]
\centering
\caption{Performance (\%) on high-risk ATCo--pilot communication pairs.}
\label{tab:safety_critical}
\begin{tabular}{llccc}
\toprule
Category & Communication Type & Acc & F1 & Recall \\
\midrule
\multirow{2}{*}{Ground Operations} 
& Line-up Clearance & 93.18 & 91.10 & 93.18 \\
& Takeoff Clearance & 93.33 & 93.24 & 93.33 \\
\midrule
\multirow{3}{*}{Airborne Control} 
& Altitude Change & 95.50 & 95.51 & 95.50 \\
& Heading Change & 100.00 & 100.00 & 100.00 \\
& Offset Instruction & 100.00 & 100.00 & 100.00 \\
\midrule
\multirow{2}{*}{Operational Coordination} 
& Contact Instruction & 97.22 & 98.15 & 97.22 \\
& Frequency Transfer & 95.05 & 95.03 & 95.05 \\
\bottomrule
\end{tabular}
\end{table}

As shown in \hyperref[tab:safety_critical]{Table 9}, SCOPE maintains consistently strong performance across high-risk ATCo--pilot communication pairs. The framework achieves over 93\% Acc and Recall on runway-related communications and reaches 95.50\% Acc on altitude change communications. It also performs reliably on operational coordination, achieving 97.22\% Acc for contact instructions and 95.05\% for frequency transfer instructions. These results demonstrate that SCOPE can provide highly accurate anomaly detection in the ATC domain.

\begin{figure}[h!]
    \centering
    \includegraphics[width=0.9\linewidth]{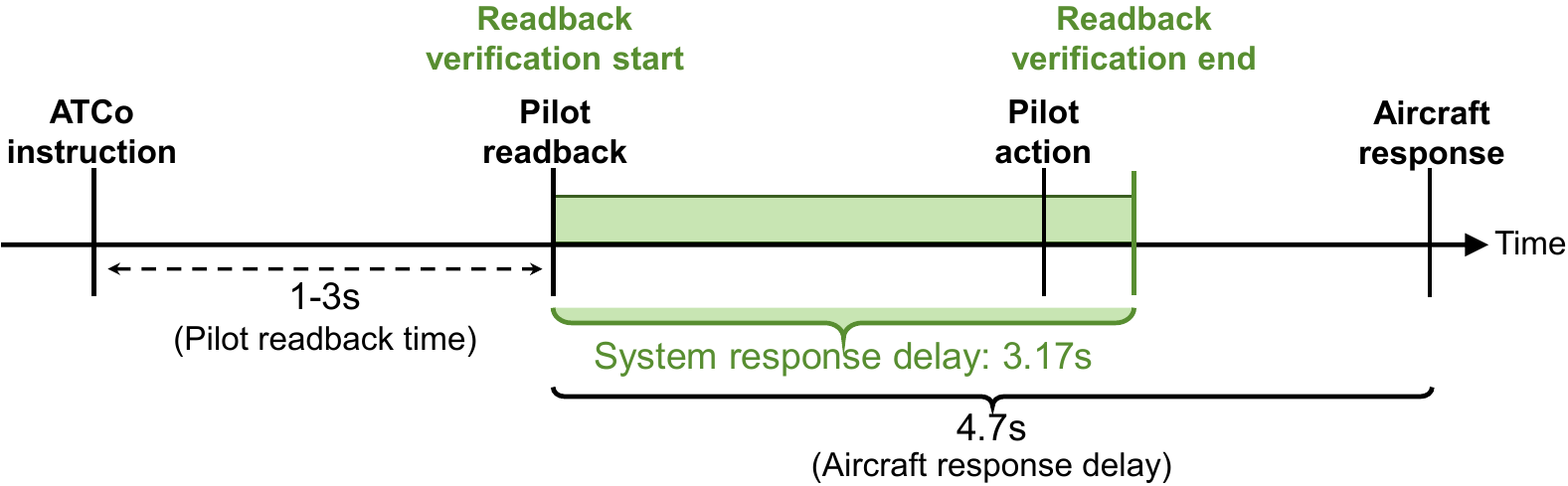}
    \caption{System response latency in relation to the ATC communication and control loop.}
    \label{fig:system response}
\end{figure}

We further analyze the response latency of the proposed system in relation to the real ATC communication and control loop, as shown in \hyperref[fig:system response]{Fig. 16}. In this workflow, the pilot readback time is approximately 1 to 3 seconds after instruction issuance~\citep{cardosi1991analysis}, and the subsequent aircraft response delay is characterized by an average response time of 4.7 seconds according to the ICAO pilot response model~\citep{icao2021acas}. Although the full SCOPE model using Qwen3-14B introduces reasoning overhead, we adopt Qwen3-4B as the backbone to balance accuracy and efficiency. Under this configuration, the proposed system achieves an average inference time of 3.17 seconds per sample, while maintaining 88.07\% accuracy and 88.17\% F1 score. Given that the available operational response window includes both the readback interval and the subsequent action response delay, the proposed system has the potential to provide timely alarms before the aircraft fully responds to the issued clearance.

\section{Conclusion}
\label{sec:conclusion}
In this paper, we propose SCOPE, a lightweight-training LLM framework for open-set ATCo--pilot readback monitoring under ICL. First, through independent class probability modeling and KNN geometric detection, the POC module provides prior readback labels, allowing the LLM to leverage the classification strength of a lightweight model for unseen communication detection. Second, the DEAR module exploits the asymmetry between ATCo instructions and pilot readbacks by anchoring retrieval on instruction relevance and preserving diversity across readback examples, thereby selecting representative demonstrations. Third, the ATCoT module performs structured semantic reasoning over intent and slot annotations, enabling the LLM to refine the plug-in label into a more accurate prediction. After anomaly detection, SCOPE generates an interpretable explanation that describes the semantic evidence behind the decision. A correction module then generates a corrected readback aligned with standard radiotelephony phraseology. On the APCP dataset, SCOPE achieves 91.05\% accuracy and 91.01\% F1 using a local unmodified Qwen3-14B under the four example setting without any LLM training. It outperforms the state-of-the-art baselines by 4.03\% in accuracy and 4.22\% in F1.
Moreover, the synergy analysis between the plug-in and the LLM shows that SCOPE can correct errors that the plug-in model fails to resolve, especially for subtle anomalous readbacks.
For explanation and correction, LLM-as-a-Judge evaluations from commercial frontier models report explanation accuracy scores above 4.6 out of 5, while the correction module achieves an overall correction rate of 96.63\%.
We further evaluate SCOPE on high-risk communication pairs. The framework achieves 100\% accuracy and recall in both heading change and offset instruction scenarios. Meanwhile, the smaller 4B variant supports efficient and timely deployment in real operational workflows.

In future work, we plan to extend SCOPE in two directions. First, we will integrate complementary operational data sources, such as radar trajectories and flight plans, to support more accurate early warning for ATCo--pilot communication anomalies. Second, we will embed the framework into real ATC monitoring and flight control environments to evaluate its human-machine collaboration capability under live operational conditions.

\section*{Acknowledgement}
This work was supported by the National Natural Science Foundation of China under Project No. 52572349, and the Hong Kong University of Science and Technology (HKUST) through the IRS Grant under Project No. IRS26EG02. The opinions, findings, conclusions, and recommendations expressed in this material are solely those of the authors and do not necessarily reflect the views or policies of the project sponsors.

\newpage

\bibliographystyle{model5-names}\biboptions{authoryear}
\bibliography{main.bib}

\end{document}